%% file: main.tex
\documentclass[acmsmall]{acmart}

\AtBeginDocument{%
  \providecommand\BibTeX{{%
    \normalfont B\kern-0.5em{\scshape i\kern-0.25em b}\kern-0.8em\TeX}}}

\setcopyright{acmcopyright}

\acmConference{arXiv.org}
\acmBooktitle{v4}

\usepackage{multirow}
\usepackage{bbding}
\usepackage{subfigure}

\newcommand{\y}{\Checkmark}
\newcommand{\n}{\XSolidBrush}

\newcommand{\mrow}[2][c]{\begin{tabular}{@{}#1@{}}#2\end{tabular}}

\def\production{1} 

\if\production0
    \usepackage{todonotes}
    \usepackage{xcolor}
    \usepackage[normalem]{ulem}
    
    \newcommand{\del}[2][black]{{\color{#1}\sout{#2}}}
\else
    
    \newcommand{\del}[2][black]{}
\fi

\begin{document}

\title[The OARF Benchmark Suite]{The OARF Benchmark Suite: Characterization and Implications for Federated Learning Systems}

\author{Sixu Hu}
\email{sixuhu@comp.nus.edu.sg}
\orcid{0000-0001-5048-3707}
\affiliation{%
    \institution{National University of Singapore}
    \streetaddress{NUS School of Computing, COM1, 13, Computing Dr}
    \country{Singapore}
    \postcode{138601}
}

\author{Yuan Li}
\email{liyuan@comp.nus.edu.sg}
\authornote{Both authors contributed equally to this research.}
\author{Xu Liu}
\email{liuxu@comp.nus.edu.sg}
\authornotemark[1]
\affiliation{%
    \institution{National University of Singapore}
    \streetaddress{NUS School of Computing, COM1, 13, Computing Dr}
    \country{Singapore}
    \postcode{117417}
}

\author{Qinbin Li}
\email{qinbin@comp.nus.edu.sg}
\affiliation{%
    \institution{National University of Singapore}
    \streetaddress{NUS School of Computing, COM1, 13, Computing Dr}
    \country{Singapore}
    \postcode{117417}
}

\author{Zhaomin Wu}
\email{zhaomin@comp.nus.edu.sg}
\affiliation{%
    \institution{National University of Singapore}
    \streetaddress{NUS School of Computing, COM1, 13, Computing Dr}
    \country{Singapore}
    \postcode{117417}
}

\author{Bingsheng He}
\email{hebs@comp.nus.edu.sg}
\affiliation{%
    \institution{National University of Singapore}
    \streetaddress{NUS School of Computing, COM1, 13, Computing Dr}
    \country{Singapore}
    \postcode{117417}
}

\renewcommand{\shortauthors}{Hu, et al.}

\begin{abstract}
    This paper presents and characterizes an Open Application Repository for Federated Learning (OARF), a benchmark suite for federated machine learning systems. Previously available benchmarks for federated learning have focused mainly on synthetic datasets and use a limited number of applications. OARF mimics more realistic application scenarios with publicly available data sets as different data silos in image, text and structured data. Our characterization shows that the benchmark suite is diverse in data size, distribution, feature distribution and learning task complexity. The extensive evaluations with reference implementations show the future research opportunities for important aspects of federated learning systems. We have developed reference implementations, and evaluated the important aspects of federated learning, including model accuracy, communication cost, throughput and convergence time. Through these evaluations, we discovered some interesting findings such as federated learning can effectively increase end-to-end throughput. The code of OARF is publicly available on GitHub\footnote{\url{https://github.com/Xtra-Computing/OARF}}.
\end{abstract}



\maketitle
\input{sections/1-introduction.tex}
\input{sections/2-backgrounds.tex}
\input{sections/3-related-works.tex}
\input{sections/4-oarf.tex}
\input{sections/5-experiments.tex}
\input{sections/6-conclusions.tex}

\begin{acks}
This research is supported by the National Research Foundation, Singapore under its AI Singapore Programme (AISG Award No: AISG2-RP-2020-018). Any opinions, findings and conclusions or recommendations expressed in this material are those of the authors and do not reflect the views of National Research Foundation, Singapore.
\end{acks}

\bibliographystyle{ACM-Reference-Format}
\bibliography{citations}

\pagebreak
\appendix
\input{sections/11-appendix.tex}

\end{document}

%% file: sections/1-introduction.tex
\section{Introduction}\label{sec:introduction}

Federated Learning (FL), first introduced by \citet{pmlr-v54-mcmahan17a}, is a technique that enables multiple parties to train a model collaboratively without leaking their private data. Recently, FL has become a heated research topic in both industry and academia. Various machine learning models, communication methods, privacy-preserving methods and data splitting schemes have been researched under federated settings.

Despite the success in those research and developments, their prosperity brings issues: There lacks a uniform standard to compare between them. Most of the studies focusing on a single aspect of FL use their own suite of datasets, models, training method and metrics. For example, previous studies on the impact of not independent and identically distributed (non-i.i.d) datasets~\cite{zhao_federated_2018,sattler_robust_2020,li_federated_2020-1} are using different data partitioning method and datasets, and previous studies on mitigating the communication cost~\cite{wang_cmfl_2019,beguier_safer_2020,li_ggs_2020,sattler_robust_2020} use different metrics to measure the speedup. Thus, a new benchmark system is needed to study and compare various FL designs, and help guide the design and implementation of future FL systems.

Looking back in history, benchmarks have played an important role in the machine learning area. Benchmark suites like DAWNBench~\cite{coleman2017dawnbench} and TBD~\cite{zhu_benchmarking_2018} have provided various benchmark metrics and results for deep learning training and inference. These benchmarks facilitate the comparison between machine learning frameworks and models. However, compared to the machine learning area, there lacks a comprehensive federated learning benchmark and a corresponding system as a reference for researchers and developers.

There already exist some preliminary benchmarks~\cite{caldas_leaf_2018,nilsson_performance_2018,luo_real-world_2019,liu_evaluation_2020} targeting FL applications. However, these benchmarks expose deficiencies. One of the most prominent is the comprehensiveness of the benchmarks is not sufficient for research needs. Another deficiency is that many of them only include non-federated datasets, where datasets are artificially split from a single dataset, which can be unrealistic scenarios. The types of models and metrics they have considered are also limited. For example, the benchmark by \citet{nilsson_performance_2018} provides a performance evaluation method based on Bayesian correlated t-test, but it only focuses on how communication architecture and data distribution affect the model accuracy.

A good benchmark for FL systems needs to address a series of factors. First, it needs to reflect the real-world scenarios by providing dedicated datasets and workloads. Second, it needs to include various metrics and measuring methods to depict a full picture of FL systems. Recent surveys on federated learning~\cite{li_survey_2021} suggested that in addition to the issues that already exist in the machine learning area, there are at least data partitioning, privacy, communication, fairness issues that need to be identified.

Taking these factors into consideration, we present OARF, a benchmark suite that aims to evaluate and study the properties of different FL systems, and provides tools to help design next-generation FL platforms. There are many open problems in the important aspects of federated learning systems, including model accuracy, communication cost, differential privacy, secure multiparty computation and vertical federated learning. OARF ensembles diversified workloads to \emph{quantitatively} assess each of those aspects. Our work highlights the following contributions.

\begin{itemize}
    \item \emph{Unified methods for measuring different aspects of FL systems.} Like any other benchmark systems, the most fundamental contribution of our work is to provide a standardized protocol and facilitates the comparison between different works. The standardization includes the identification of major components of federated learning, the methods for measuring each of those components and their combinations, and the metrics to measure the results.
    \item \emph{Publicly available datasets for simulating data silos for meaningful application scenarios.} Like machine learning models, the quality of FL models are heavily dependent on the quality of datasets. Most FL benchmarks split a single machine learning dataset and distributed the subsets to different parties for federated training, which is convenient in practice but sometimes masks the source of accuracy impacts on FL models. It is hard to decide whether those impacts come from the federation, or instead, come from the homogeneity of the subsets. In real-world scenarios, the datasets are often from different parties and are heterogeneous. We have collected and assembled real-world datasets from different sources and designed workloads covering numerous domains to reflect this situation. We believe that this process itself is a non-trivial task, after examining multiple datasets according to our design goals.
    \item \emph{Quantitative study on the intrinsic properties of federated learning and its components.} Through reference implementations and preliminary experiments, we quantitatively study the intrinsic properties of FL systems, including the preliminary relations between various design metrics such as data partitioning, privacy mechanisms and machine learning models. These properties enrich our knowledge of the internal mechanism for federated learning, provide valuable experience for the industry needs, and provide suggestions on building future FL frameworks.
    \item \emph{General framework and reference implementations that show the future research opportunities for important aspects of federated learning systems.} Reference implementations are provided in our work for better reproducibility and modularity. Each reference implementation evaluates one or more properties stated above. We conclude the common evaluation results of these implementations, including the performance boost brought by FL and the time decomposition of FL training process, as our findings and research opportunities for future FL systems.
\end{itemize}

In the rest part of this paper, Section~\ref{sec:backgrounds} and~\ref{sec:related_works} introduce the basics of federated learning and existing works on the benchmarks. Section~\ref{sec:oarf} first presents the design principles of the OARF benchmark, and how these principles help solve the challenges in FL benchmarks. Then it describes OARF in detail, including the datasets, workloads and details of our reference implementations. Following that, Section~\ref{sec:experiment} presents the preliminary experiment results to answer a series of questions we encountered in FL system implementations. Finally, Section~\ref{sec:conclusion} concludes our studies and give insights for future FL frameworks and applications.

%% file: sections/2-backgrounds.tex
\section{Backgrounds}\label{sec:backgrounds}

One of the major challenges of creating a comprehensive benchmark suite lies in the complexity of federated learning itself. After introduced by~\citet{pmlr-v54-mcmahan17a}, the content of federated learning was enriched extensively. Recent surveys by~\citet{li_survey_2021} and~\citet{yang_federated_2019} indicate that FL is a combination of various techniques in multiple areas, and designing a comprehensive benchmark that covers all these aspects, including their performance and their mutual effects is inherently challenging.

    As a benchmark, our paper differs significantly from those survey papers by providing a quantitative approach for evaluating existing FL systems and compare between them. Existing survey papers do not provide such quantitative metrics, detailed FL system architecture and reference implementations.

    Inspired by these papers, our work performs a further step, using detailed experiments to quantitatively evaluate the characteristics of each of those aspects. In such way we try to provide valuable references and design guides for future application developments on FL system. FL is still in its early and fast developing stage, and we do not see a dominant or standard workload in this area (such as the ImageNet in the computer vision area). That is why our decision is to focus on the breadth. Out of the same reason, we have to implement our own workloads and try to cover as many aspects as possible. On the other hand, although focusing on breadth, depth can still be pursued for each workload in the benchmark, as demonstrated by the extensive studies on two workloads in our experiments.

\subsection{Data Partitioning}\label{ssub:data_partitioning}

Data partitioning is one of the most fundamental features of federated learning. It not only describes the relation of each party's datasets but also restricts which federated algorithms can be applied. \citet{yang_federated_2019} introduced three partition schemes: horizontal, vertical, and hybrid, where horizontal FL uses datasets that share feature space but not sample space, vertical FL uses datasets that share sample space but different in feature space, and hybrid FL uses datasets that are different in both sample and feature spaces.

One scenario for the horizontal partitioning scheme is that two or more different companies of the same type have different datasets of users that share similar features for each user, and wish to study the common behavior of the users from their datasets. Numerous recent works on FL are based on the horizontal partition scheme. For example, Google in 2017 proposed a horizontal federated learning model to improve the google keyboard suggestion quality~\cite{yang_applied_2018} and later generalized their method to a general federated learning system~\cite{bonawitz_towards_2019}.

The Horizontal partitioning scheme also brings an interesting problem: To mitigate the impact brought by the difference of data distributions across parties. \citet{zhao_federated_2018} illustrated how non-i.i.d-ness of data can affect the global model's performance, while \citet{hsieh_non-iid_2020} and \citet{kairouz_advances_2021} formally classified the non-i.i.d-ness into 5 classes: feature distribution skew, label distribution skew, two types of concept shift, and quantity skew, which provides clues to quantitatively explore their impacts.

For a benchmark system, however, as most of the known works use their own partitioning methods and measuring metrics, it is challenging to provide a uniform way for obtaining such non-i.i.d datasets for training, either by partitioning a single dataset or by combining multiple different detests together, and quantitatively measure their non-i.i.d-ness.

For vertical FL, as it requires a shared sample space, it is unlikely to find a large number of datasets that contain the same set of identities. Thus, scaling-up vertical FL is difficult. Recent works, regardless whether it is linear regression~\cite{Gascn2016SecureLR}, tree-based system such as SecureBoost~\cite{cheng_secureboost_2021} and Federated Forest~\cite{liu_federated_2020}, or neural networks~\cite{gu_privacy-preserving_2021}, are using either a limited number of parties, or synthetic datasets for the experiments.

The datasets used for vertical FL often require \emph{entity alignment} techniques~\cite{christen_data_2012} to find the entries that can be combined together. Apart from synthetic datasets, most public datasets targeting the same kind of tasks are suitable for horizontal-partitioned FL, but only a few datasets that embed the concept of ``identity'' can be adopted to the vertically-partitioned scenario, thus finding such datasets and regulating the entity alignment scheme are essential for a benchmark system.

In the case of hybrid data partitioning, two datasets overlap in both feature and sample space, but none of the spaces is identical. Transfer learning~\cite{pan_survey_2010} might be used to make up the inconsistency in feature and sample spaces between the two datasets. \citet{liu_secure_2020} present a successful example of using federated transfer learning.

\subsection{Federated Learning Algorithms}\label{ssub:fl_algorithms}
FL algorithm defines the training process --- a protocol describing how each party performs their training and how information from different parties is aggregated into a single model. The data partitioning scheme limits the range of FL algorithms that can be applied.

The horizontal FL algorithm can be classified into two categories: centralized and decentralized~\cite{li_survey_2021}. In the centralized design, there exists a central party to aggregate each party's information and update the global model. Federated averaging (FedAvg)~\cite{pmlr-v54-mcmahan17a} is perhaps the most widely used one of them. There exists other algorithms with centralized design such as Federated Stochastic Variance Reduced Gradient (FedSVRG)~\cite{konecny_federated_2016}, Federated LSTM (FedLSTM)~\cite{mcmahan_learning_2018}, Distributed Selective SGD (DSSGD)~\cite{shokri_privacy-preserving_2015}, Federated Proximal (FedProx)~\cite{li_federated_2020-1}, Federated Normalized Averaging (FedNova)~\cite{wang_tackling_2020}, etc. Each of them targets a specific issues such as data distribution, underlying models, and privacy, but most of them are derived from or resembles the FedAvg algorithm.

The decentralized algorithms are not well studied as the centralized ones, due to their being harder to design and implement. The key idea of them is to replace the central node with a peer-to-peer (P2P) protocol. Such a design is helpful to mitigate overloaded servers, single-point failures, and privacy issues. Several works are focusing on the overcoming the data variance challenges~\cite{tang_d2_2018}, topology challenges~\cite{yu_distributed_2019}, communication challenges regarding network capacities~\cite{hu_decentralized_2019}, trust~\cite{he_central_2019}, privacy~\cite{bellet_personalized_2018}, and algorithm design for tree-based systems~\cite{li_practical_2020}. But there lacks a consensus on the universal design of efficient decentralized learning architecture.

Most vertical FL algorithms adopt the centralized design. As the input features are of different shapes for each party, the models of different parties are heterogeneous, so the decentralized design cannot easily be adopted. Split learning~\cite{vepakomma_split_2018} is proposed as a baseline solution for this problem. In the split learning setting, each client maintains a part of the neural networks, from the input layer to the \textit{cut layer}. In a training iteration, each client calculates its own forward pass until the cut layer, then sent the parameters of the cut layer to the server. The server then concatenates all clients' data and continues the rest forward pass until the output layer. Then gradient is then back-propagated in a similar way.

Although split learning solves the architectural problem, it becomes cumbersome when security and privacy are involved in this process. The communicated data can leak information about the input data. GELU-net~\cite{zhang_gelu-net_2018} provides a strong encryption method to solve this issue using homomorphic encryption, but their methods only encrypt the communicated data, and the plain-text server parameters expose possible vulnerabilities to leak information. NoPeek SplitNN~\cite{vepakomma_nopeek_2020} provides another idea to reduce privacy leakage by minimizing the distance correlation between raw data, but this method does not eliminate the chance of reconstructing the original data. Providing a strong and efficient privacy guarantee in the split learning setting remains an open problem.

\subsection{Underlying Models}\label{ssub:underlying_models}

Federated learning requires an underlying machine learning model to work on actual tasks, and the selection of the model affects how federated learning is implemented. For example, a number of existing works~\cite{wang_cmfl_2019,pmlr-v54-mcmahan17a,Wang2020Federated} use neural networks as their training models, and federated stochastic gradient descent based methods are widely applied on neural networks in these works. Other models like decision trees~\cite{li_practical_2020} and linear models~\cite{hardy_private_2017} could also be used as the underlying models. However, when using these types of models, techniques such as federated stochastic gradient descent may no longer be suitable. Taking the SecureBoost model~\cite{cheng_secureboost_2021} as an example, it is natural to vertically split the dataset by features, distribute the sub-datasets to different parties, and use a tree-based structure to organize each party's sub-model and combine them as a large one. But for various neural network models, there is no such obvious method for training vertically-partitioned datasets. In this paper, we only focus on neural networks due to their wider application range and more extensive research in the federated learning area.

\subsection{Data Security and Privacy Protection}\label{ssub:secure_multiparty_computation}

One key reason for preferring federated learning over training all data centrally in one machine is to protect each party's privacy, or more specifically, prevent information from leaking in the training and evaluation process. Various methods are used to achieve this goal, in which the most prominent one can be roughly categorized into cryptographic methods and differential privacy.

Cryptographic methods such as \emph{secure multi-party computation} (SMC)~\cite{yao_protocols_1982} and \emph{homomorphic encryption} (HE)~\cite{gentry_fully_2009} are widely applied to machine learning~\cite{du_building_2002,wan_privacy-preservation_2007,phong_privacy-preserving_2018}, and they ensure that the information is not leaked when transferring data between different parties. The goal of both of them is to compute a function \(f\) on private data \(x_{1}, x_{2},\dots,x_{N}\) which is scattered on \(N\) different parties, and produce \(f(x_{1},x_{2},\dots,x_{n})\) while keep each parties' data secret, but they adopted different approaches. SMC utilizes secure protocols such as oblivious transfer~\cite{lo_insecurity_1997} to ensure that each the party that calculates \(f\) does not know the original message, while HE exploits the homomorphism of encryption functions, encrypting each pieces of data before performing the calculation, and decrypting the calculated result afterwards.

Numerous methods have been proposed to implement SMC, among which the most practical ones for SMC are Garbled Circuit~\cite{yao_how_1986} and SPDZ~\cite{safavi-naini_multiparty_2012}, and for HE there are schemes based on cryptosystems such as Lattice~\cite{gentry_fully_2009} and Paillier~\cite{stern_public-key_1999}. But these protocols and schemes have common drawbacks for both FL frameworks and their benchmark systems: they introduce excessive time and space overhead. For instance, the Paillier HE introduces over 40\(\times\) data expansion and is extremely slow to calculate~\cite{yang_batchcrypt_2019}, thus for a benchmark, finding a suitable algorithm that is simple to apply and has a reasonable amount of overhead is essential. Quantifying the overhead of such a method will provide both baselines for future research and insights for developing new security mechanisms.

Differential privacy, on the other hand, protects individuals' privacy in the released model instead of data security in the communication process. Please refer Appendix~\ref{ssub:differential_privacy} to for the detailed mechanisms of differential privacy.

\subsection{Communication Cost}\label{ssub:bg_communication_cost}

Apart from the security and privacy, another major concern unique to FL is the communication costs, due to the distributed nature of federated learning and their impacts on model performance. A large amount of communication is required between each party in the training process of federated learning regardless of the communication architecture. This can become a large problem or even a primary bottleneck when the size of the underlying model or the number of parties is large. Methods proposed to mitigate this problem can be roughly categorized into high-level methods and low-level methods.

High-level methods, such as CMFL~\cite{ribero_communication-efficient_2020} or Optimal client sampling~\cite{wang_cmfl_2019}, try to reduce the amount of communication data by only selecting the updates that are likely to improve the global model. These kinds of methods do not change the structure of the data sent between each party, but only reduce the total number of messages packets sent.

Low-level methods, such as deep gradient compression (DGC)~\cite{lin_deep_2018}, QSGD~\cite{NIPS2017_6768}, GGS~\cite{li_ggs_2020} and SketchML~\cite{jiang_sketchml_2018}, tries to compress the gradient or model parameters sent between each party, often by sub-sampling, quantization and sketching. Although these low-level methods often achieve higher compression ratios than the high-level methods, different methods often achieve different maximum compression ratios, and there lacks a uniformity between them.

For a benchmark system, evaluating and comparing these methods needs a non-trivial design since the original works are using different methods and metrics to measure the compression ratio and the impact on model performance. There is also no uniformity between high-level methods and low-level methods.

%% file: sections/3-related-works.tex
\section{Related Works}\label{sec:related_works}

There have been some efforts for benchmarking FL systems. Table~\ref{tab:benchmark-comparison} summarizes the characteristics of these benchmarks and related works. Following \citet{li_survey_2021}, these benchmarks can be categorized into three categories: datasets, targeted benchmarks, and general-purpose benchmarks. To the best of our knowledge, no benchmark is comprehensive enough to evaluate all the important aspects of FL described in Section~\ref{sec:backgrounds}.

\begin{table}[htb]
    \centering
    \caption{Features of FL Benchmarks}\label{tab:benchmark-comparison}
    \begin{tabular}{c l c c c c c}
        \toprule
        Category &
        Name     &
        \mrow{Federated \\Dataset}       &
        \mrow{Partitioning \\Scheme}     &
        \mrow{Various \\Models}       &
        \mrow{Prov./Sec. \\Mechanism}   &
        \mrow{Comm. \\Cost} \\
        \cmidrule(lr){1-7}

        \multirow{2}{*}{Datasets}
                 & LEAF~\cite{caldas_leaf_2018}            & \n & \n & \y & \y & \y \\
                 & Street Image~\cite{luo_real-world_2019} & \n & \n & \y & \n & \y \\
        \cmidrule(lr){1-7}
        \multirow{6}{*}[-1ex]{\mrow{Targeted \\Benchmarks}}
                 & \citet{nilsson_performance_2018}        & \n & \n & \n & \n & \y \\
                 & \citet{liu_revocable_2019}              & \y & \n & \y & \y & \n \\ 
                 & \citet{liu_evaluation_2020}             & \n & \n & \n & \n & \y \\
                 & \citet{gao_end--end_2020}               & \n & \y & \n & \n & \y \\
                 & \citet{zhang_improving_2021}      & \n & \n & \n & \n & \y \\
                 & \citet{zhuang_performance_2020}         & \y & \n & \y & \y & \n \\
        \cmidrule(lr){1-7}
        \multirow{5}{*}[-1ex]{\mrow{General \\Purpose\\Benchmarks}}
                 & Edge AIBench~\cite{hao_edge_2019}       & \n & \n & \y & \n & \n \\
                 & FLBench~\cite{gao_flbench_2021}    & \y & \n & \y & \y & \y\\
                 & FedML~\cite{he_fedml_2020}              & \n & \y & \y & \n & \n \\
                 & FedEval~\cite{chai_fedeval_2020}        & \n & \y & \y & \y & \y \\
                 & OARF~(our work)                         & \y & \y & \y & \y & \y \\
        \bottomrule
    \end{tabular}
\end{table}

\subsection{Datasets}\label{ssub:datasets}

These types of works aim at providing datasets dedicated to federated learning. Unlike machine learning, FL asks for specific requirements on datasets, including the number of parties, data distribution, and versatility of configuring datasets. Solving these requirements is the main purpose of these types of works.

LEAF~\cite{caldas_leaf_2018} provides several image/text datasets and a set of reference implementations using federated averaging. It also includes various measuring metrics to evaluate accuracy performance and computational cost in edge devices under different settings. However, LEAF only focuses on the massively cross-device scenario, where federated learning is performed on a massive number of devices, but neglects the cross-silo scenarios.

\citet{luo_real-world_2019} proposes a street image dataset, which provides high-quality labeled data for FL research. They evaluate the accuracy and communication costs of YOLO and Faster R-CNN under different settings. Their paper focuses more on the properties of the proposed dataset instead of the federated learning system.

In comparison, a unique feature of OARF is that OARF assembles multiple real-world datasets to create more realistic application scenarios, where the previous studies mainly utilize data partitioning on a single dataset for creating the data silo scenarios.

\subsection{Targeted Benchmarks}\label{ssub:targeted_benchmarks}

Targeted benchmarks evaluate and try to optimize aspects in a small domain. These types of benchmarks often provide valuable information in their focusing domain but are also limited by that domain when a more comprehensive view of an entire FL system is required.

\citet{nilsson_performance_2018} have proposed a benchmark to measure the performance of three different FL algorithms, namely Federated Averaging (FedAvg)~\cite{pmlr-v54-mcmahan17a}, Federated Stochastic Variance Reduced Gradient (FSVRG)~\cite{konecny_federated_2016}, and CO-OP~\cite{wang_co-op_2017}. Their work uses the Bayesian t-test to compare two algorithms. However, this benchmark is designed only for comparing the accuracy performance of two FL algorithms, but not learning other characteristics of FL systems, such as privacy mechanisms and communication cost.

\citet{liu_revocable_2019} implement a benchmark targeting federated forest. Their works focus on the revocation problem where participant tries to protect their data exposed in the main model after they quit the training process. The experiment result shows that the designed protocol can ensure that the information of a revoked participant is securely removed.

\citet{liu_evaluation_2020} introduce an evaluation framework for a large-scale benchmark that focuses on exploring the relationship between the skews of datasets. It discusses the effects of datasets properties, including quantity distribution skew, label distribution skew, in FL systems. However, it lacks the discussion of other components in FL systems, such as privacy and communication architecture.

\citet{gao_end--end_2020} study horizontal FL and split neural networks (SplitNN) on IoT devices. Multiple metrics regarding time, communication, memory, and power usage are taken into consideration. Their results show that FL performs slightly better than SplitNN due to smaller communication overhead, and neither FL nor SplitNN is suitable for large models due to limited hardware resources in IoT platforms.

\citet{zhang_improving_2021} developed a benchmark system in semi-supervised FL settings, where clients' data are unlabeled and the quantity of data on the server is limited. Multiple factors affecting the final model accuracy have been studied, such as data distributions, training setups, and amount of clients. The improved method based on the benchmark result also achieves better generalization ability, proving the effectiveness of the benchmark result.

\citet{zhuang_performance_2020} improve federated person re-identification with their benchmark results. The datasets in the benchmark are sourced from various domains to simulate real-world scenarios. The proposed optimization methods improve both convergence and performance on the federated partial averaging algorithm.

OARF is different from the targeted benchmarks above as it provides a modular and extensible design to provide references for a variety of aspects in FL, instead of focusing on just a small domain.

\subsection{General-Purpose Benchmarks}\label{ssub:general_purpose_benchmarks}

General-purpose benchmarks evaluate FL systems comprehensively and characterize different aspects of FL systems in detail, which is also the goal of our benchmark suite.

Edge AIBench~\cite{hao_edge_2019} is a test platform for FL applications. It proposes four scenarios in four domains: medical, surveillance, home living, and vehicle, as references. However, no experiment result is available at the time of writing this paper. This platform is listed here out of completeness consideration, as it mainly focuses on the edge computing FL scenarios, while our paper focuses on the cross-silo scenarios.

FLBench~\cite{gao_flbench_2021} is a benchmark suite covering multiple domains. It measures aspects such as communication, privacy, data distribution, and various training settings. It also features configurable scenarios and automated deployment, but like Edge AIBench, it is still under development currently and no experiment result has been reported.

FedML~\cite{he_fedml_2020} serves as both an FL framework and a benchmark suite. It implements numerous practical FL algorithms in both horizontal and vertical FL settings and is capable of running these algorithms in different communication settings such as distributed training and standalone simulation. It is one of the most comprehensive frameworks, despite part of its evaluations to the references implementations is still preliminary.

FedEval~\cite{chai_fedeval_2020} is a FL evaluation model featuring the ``ACTPR'' (accuracy, communication, time consumption, privacy, and robustness) model. It uses virtualization techniques to isolate evaluation environments and also to bypass hardware resource limitations. Two FL algorithms, FedSGD and FedAvg, are implemented and evaluated currently. Their results show that FedAvg is more efficient regarding time and communication, but suffers more from non-i.i.d data distributions.

\subsection{Summary}\label{ssub:summary}

Summarizing existing benchmarks, they play an important role in FL research. Various characteristics can be quantitatively analyzed with different types of targeted and general-purpose benchmarks, and the dataset research provides significant infrastructures.

Currently, there exists no comprehensive enough benchmark system that covers all the aspects of FL systems. Even the most comprehensive ones show deficiencies of algorithms and metrics on different granularities of the system. This is an especially prominent issue in terms of security and privacy. Their measurement and analysis in FL systems are still preliminary and require further development.

Moreover, the consensus issue between different studies also needs further efforts. Taking FL dataset usage as an example, most research use synthetic datasets split from a single source, while they are using different methods to perform the splitting. Similarly, no consensus has been established on the metric of non-i.i.d-ness. Although as proposed in FedML~\cite{he_fedml_2020}, using a realistic partitioning method can mitigate this issue, realistic partitioning is not suitable for large-scale FL, since collecting data from different sources is inherently difficult.

%% file: sections/4-oarf.tex
\section{The Benchmark Suite Design}\label{sec:oarf}

To complement deficiencies of existing benchmark studies in FL, our work focuses on:
\begin{itemize}
    \item Provide to researchers and developers a modular, extensible, and accessible FL framework.
    \item Provide combinable datasets collected from different sources and corresponding manipulation tools to reflect real-world scenarios.
\end{itemize}

\subsection{Design Principles}

According to Jim Gray's benchmark handbook~\cite{gray_benchmark_1994}, a good benchmark system should have four characteristics: \textit{relevant}, \textit{portable}, \textit{scalable} and \textit{simple}. The former three principles are ensured by repeated experiments, flexible and modular design of our system, and providing scalable datasets and data splitting tools. Fulfilling the simplicity principle makes us decouple each aspect into a single module and analyze them separately. Using these methods we can not only combine arbitrarily different modules to form complex systems, but also make sure that the experiment environment is easily reproducible and the time required for most benchmark targets is within an acceptable range, benefiting both users and contributors.

The OARF benchmark suite not only achieves these four goals but also embeds other good features in the design, as explained below. These features serve both as complements and enticements to the four goals.

\begin{itemize}
    \item \emph{Diversity of Workloads.} Federated learning can be used in several domains, such as biology, finance, and mobile applications. Tasks in different domains vary greatly from each other. Data formats the machine learning models vary from task to task. For different tasks, the data distribution and data partitioning scheme are also different. Although it is impractical to design a benchmark to cover all kinds of tasks, efforts are made in OARF to maximize the diversity of workload, so that different types of tasks and the corresponding applications can be characterized.
    \item \emph{Comprehensiveness.} Federated learning can be characterized in various ways. As presented in Section~\ref{sec:backgrounds}, it can be at least categorized in six aspects, including data partitioning, FL algorithms, underlying models, security, privacy, and communication cost. To cover all these aspects, OARF uses various metrics to measure the performance of each component in federated learning, and explore how they contribute to the entire learning process.
    \item \emph{Openness.} Since federated learning is a rapidly developing field, concepts may be added into FL systems in the future, and demands for FL are rapidly developing and existing benchmarks may fall behind. Taking these into consideration, we make OARF open to modification and addition. Resulting from the modular design of our benchmark, its users can easily contribute or modify different parts of it to meet their specific requirements.
\end{itemize}

In addition to the benchmark part, the principles of design OARF as an FL framework is also taken into consideration. One of the most important guidelines is to make the whole architecture layered and modular so that it can be easily extended and modified to meet various demands.

\subsection{Overview}

\begin{figure}[htb]
    \centering
    \includegraphics[width=0.95\linewidth]{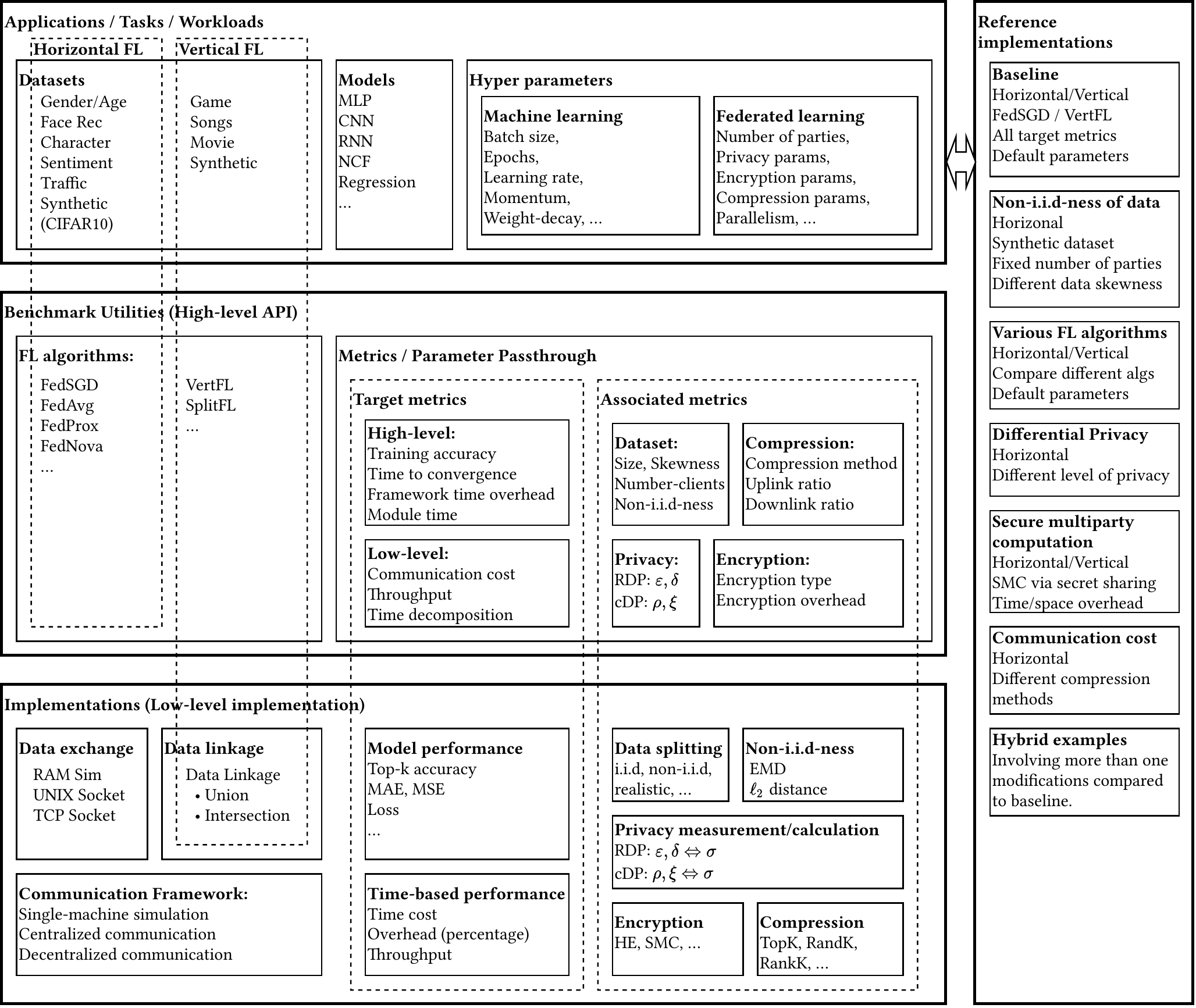}
    \caption{OARF framework architecture}\label{fig:architecture}
\end{figure}

Figure~\ref{fig:architecture} shows the architecture of the OARF benchmark. It adopts a layered design, with three layers namely applications, benchmark utilities (high-level API), and implementations (low-level API). Additionally, reference implementations utilize functionalities of all three layers, but only directly interact with the application layer.

The application layer provides an interface for end-users, allowing them to configure datasets, models, and hyperparameters according to their demands. After parsing the hyperparameters, the application layer invokes functionalities provided by the high-level API provided in the benchmark utility layer to perform the actual benchmarking.

The benchmark utility layer is divided into two parts. The FL algorithms part implements various FL algorithms such as FedAvg~\cite{pmlr-v54-mcmahan17a}, FedProx~\cite{li_federated_2020-1}, and FedNova~\cite{wang_tackling_2020}, which also takes all kinds of ``add-ons'' such as privacy, security, and compression into consideration and ensures that they work together with the FL algorithms flawlessly. This is a non-trivial task as the original work proposing these FL algorithms often do not take these aspects into consideration. The other part ``Metrics/Parameter Pass-through'' serves as a bridge to communicate between the application layer and the implementation layer, and also decides which metrics to use.

The implementation layer is responsible for all the heavy-lifting work other than FL algorithms. It provides all the infrastructures and functionalities that are needed by FL systems, such as communication infrastructure, implementations for privacy, security and compression mechanisms, data splitting mechanisms, and all kinds of statistic codes for collecting information from the training process and calculate various metrics.

Reference implementations are defined only by the datasets and models they use, and hyper-parameters of each of the modules. Having them defined purely by configuration files makes them easy to understand, replicate and extend according to their needs. For each important aspect mentioned in Section~\ref{sec:backgrounds}, we set up a reference implementation to both explore the basic characteristics of that aspect and present the way to use our system to other developers.

Each layer can be extended or enhanced accordingly. If a required feature is not implemented in this benchmark, the layered and modular design makes it easy for developers to insert code and utilizing/incorporating existing code without spending significant time incorporating the new module into the whole system.

\subsection{Metrics}\label{sub:metrics}

Metrics play a significant role in benchmarks, unifying the method to analyze different aspects in FL and modules in FL systems. Based on the major components in and the observation of existing FL systems, we divide metrics into two parts --- target metrics and associated metrics, where target metrics measure the final results that researchers concern and try to optimize, and the associated metrics measure how can they be optimized, revealing the quantitative relation between target and various configurations of FL systems.

Currently, the target metrics implemented are similar to those of machine learning benchmarks. High-level metrics, including training accuracy, time to convergence, and framework time overhead, measures the performance of the training process and the final model, while low-level metrics, including communication cost, throughput, and time decomposition, focuses on the low levels of FL systems, such as FL infrastructure, operating system, and hardware performance.

The associated metrics are more unique to FL systems, and their selections are controversial across different studies. For example, the non-i.i.d-ness of datasets can be measured by the earth-mover's distance~\cite{zhao_federated_2018}, \(\ell_{1}\)-distance-based average total variation~\cite{zhang_improving_2021}, as well as the parameter to split the synthetic dataset~\cite{li_federated_2021}. In this paper, we provide multiple choices for metrics like this to fit different needs, but only uses and recommend one kind of them in our reference implementations to maintain uniformity. For the non-i.i.d-ness of datasets, we adopt the parameter of Dirichlet's distribution~\cite{li_federated_2021} since it is easy to control and measure. For differential privacy, we adopt \((\varepsilon,\delta)\) parameter from R\'enyi differential privacy~\cite{abadi_deep_2016} out of the same reason. For gradient/model compression, we measure the downlink compression ratio and uplink compression ratio separately. The convergence speed is also measured, since the high-level compression methods may greatly affect convergence speed without affecting final model accuracy significantly. For encryption, we record the encryption type and corresponding parameters and measure the time and data overhead.

Both high-level metrics and low-level metrics are dependent on low-level implementations. If new modules are to be added to the system, new metrics might be added accordingly. The modular metric system ensures that other modules are not affected much by the new module, and new metrics can be evaluated correctly in existing FL algorithms.

\subsection{Workloads}\label{sub:datasets}

\begin{table}[htb]
    \centering
    \caption{A list of tasks and corresponding datasets. }\label{tab:tasks}
    \newcommand{\about}{\textasciitilde{}}
    \begin{tabular}{c c c l c}
        \toprule
        P\(^*\) & D\(^*\) & Task & Datasets & \# instances         \\
        \cmidrule(lr){1-5}
        {}\multirow{15}{*}[-1.5em]{\rotatebox[origin=c]{90}{Horizontal}}
        {}  &\multirow{11}{*}[-1em]{\rotatebox[origin=c]{90}{CV}}
        {}    &\multirow{4}{*}{Age Prediction}
        {}      & All-Age-Face~\cite{cheng2019exploiting}                           &\about13k  \\
        {}  & & & IMDB-WIKI (Wiki part)~\cite{rothe_deep_2018}                      &\about62k  \\
        {}  & & & APPA-REAL~\cite{agustsson_apparent_2017,clapes2018apparent}       &\about7.6k \\
        {}  & & & UTK Faces~\cite{zhang_age_2017}                                   &\about20k  \\
        \cmidrule(lr){3-5}
        {}  & &\multirow{2}{*}{Face Recognition}
        {}      & BUPT-Balancedface~\cite{wang_deep_2021,wang_mitigating_2020,Wang_2019_ICCV}         &\about1.25M\\
        {}  & & & Racial Faces in-the-Wild~\cite{wang_deep_2021,wang_mitigating_2020,Wang_2019_ICCV}  &\about40k  \\
        \cmidrule(lr){3-5}
        {}  & &\multirow{2}{*}{\mrow{Alphanumeric Character\\Recognition}}
        {}      &  Chars74K-Fnt~\cite{campos_character_2009}                        &\about3.4k  \\
        {}  & & &  Chars74K-Hnd~\cite{campos_character_2009}                        &\about63k  \\
        \cmidrule(lr){3-5}
        {}  & &\multirow{2}{*}{\mrow{Chinese Character\\Recognition}}
        {}      &  HIT-OR3C~\cite{zhou_hit-or3c_2010}                               &\about460k \\
        {}  & & &  CASIA-HWDB\(^a\)~\cite{liu_casia_2011}                           &\about3.8M \\
        \cmidrule(lr){3-5}
        {}  & &Object Classification
        {}      &  CIFAR10 (Synthetic)~\cite{krizhevsky2009learning}                &\about60k  \\
        \cmidrule(lr){2-5}
        {}  &\multirow{2}{*}{\rotatebox[origin=c]{90}{NLP}}
        {}    &\multirow{2}{*}{Sentiment Analysis}
        {}      & IMDB Movie Review~\cite{maas-EtAl:2011:ACL-HLT2011}               &\about50k  \\
        {}  & & & Amazon Movie Review (Subsampled)~\cite{mcauley_amateurs_2013}     &\about79k   \\
        \cmidrule(lr){2-5}
        {}  &\multirow{2}{*}{\rotatebox[origin=c]{90}{GIS}}
        {}    &\multirow{2}{*}{Traffic Prediction}
        {}      & METR-LA~\cite{jagadish_big_2014}                                  &\about34k  \\
        {}  & & & PEMS-BAY~\cite{li2018dcrnn_traffic}                               &\about52k  \\
        \cmidrule(lr){1-5}
        {}\multirow{7}{*}[-0.5em]{\mrow{\rotatebox[origin=c]{90}{Vertical / Hybrid}}}
        {}  & \multirow{7}{*}[-0.5em]{\rotatebox[origin=c]{90}{General ML}}
        {}    &\multirow{3}{*}{\mrow{Trend Prediction / \\Recommendation}}
        {}      & Steam Game\footnotemark{}\(^{,b}\)                                &\about17k \\
        {}  & & & IGN Rating\footnotemark{}                                         &\about18k      \\
        {}  & & & Video Game Sales\footnotemark{}                                   &\about55k      \\
        \cmidrule(lr){3-5}
        {}  & &\multirow{2}{*}{\mrow{Trend Prediction / \\Recommendation}}
        {}      & MovieLens 1M~\cite{harper_movielens_2016}                         &\about1M   \\  
        {}  & & & IMDB Movie\(^c\)~\cite{maas-EtAl:2011:ACL-HLT2011}                &\about4M   \\
        \cmidrule(lr){3-5}
        {}  & &\multirow{2}{*}{Year Prediction}
        {}      & MillionSong~\cite{Bertin-Mahieux2011}     &\about1M      \\ 
        {}  & & & Free Music Archive~\cite{fma_dataset}     &\about106k    \\
        \bottomrule
    \end{tabular}
    \footnotesize
    \flushleft
    \bigskip
    \par\(^*\)In the headers, P stands for Data Partitioning, and D for Application Domains.
    \par\(^a\)Contains 3 independently collected subsets with size 1.6M, 1.2M and 1.0 M separately.
    \par\(^b\)The Steam Game dataset only counts the number of games, and there is also players' information in this dataset.
    \par\(^c\)The IMDB Movie dataset counts only the number of unique titles.
\end{table}
\addtocounter{footnote}{-3}
\stepcounter{footnote}\footnotetext{Steam Game: \url{https://steam.internet.byu.edu/}}
\stepcounter{footnote}\footnotetext{IGN Rating: \url{https://github.com/john7obed/ign_games_of_20_years}}
\stepcounter{footnote}\footnotetext{Video Game Sales: \url{https://github.com/ashaheedq/vgchartzScrape}}

To cover various scenarios of federated learning, we collect public available datasets and categorize them by data partition schemes and tasks, as shown in Table~\ref{tab:tasks}. Different datasets belonging to the same task can be used to simulate data possessed by different parties. All these datasets are from different sources, which means that they are ``naturally'' owned by different parties instead of manually split from a single dataset. Datasets like these more precisely reflect how data is distributed in real-world federated learning tasks.

For vertical datasets, we need to align data records from each party to train them collaboratively. In real-life scenarios, the data is often aligned by a column that identifies the user. However, most public available datasets have the user column anonymized. Even for non-anonymized datasets, there is little chance that the same user in two different data shares the same user id. Here we collected datasets that align by game title, track name, and movie title separately.

In the following, we only introduce the workloads demonstrated in the experiment part of this paper. Other workloads have been implemented in our benchmark system but omitted here to avoid redundancy.

\paragraph{Object Classification}\label{par:synthetic_workloads}
Although realistic datasets can reflect the real-wold federated distribution, it faces the issue when the number of parties is large, a large amount of independent datasets in the same task is not easy to collect. However, we can split a single dataset into multiple different subsets to simulate a federated setting. In this task, we use the CIFAR10~\cite{krizhevsky2009learning} dataset, and split it into 5 parts with various splitting methods, and use ResNet-18 to classify the images.

\paragraph{Sentiment Analysis}\label{par:sentiment_analysis}
Sentiment analysis is a technique that uses natural language processing and text analysis to study and predict affective states. The emergence of machine learning has greatly propelled the studies in this area. Recent research~\cite{wang_predicting_2015} and surveys~\cite{zhang_deep_2018} have shown that LSTM is effective for this kind of task. Today, various rating websites like IMDB and Rotten Tomato provide a large amount of data for the training of sentiment analysis model. We apply federated averaging to an LSTM model, and use movie review data from Amazon and IMDB for the training. Limited by the GPU capacity and training time, we use the whole 50,000 entries in IMDB Movie Review dataset and randomly sampled 1\% from the Amazon Movie Review dataset as the training data.

\paragraph{Year Prediction}\label{par:year_prediction}
Year-Prediction is a regression task for predicting the release year of music tracks based on the audio features such as ``energy'' or ``danceability''. We use the audio features and release year labels in the MillionSong dataset (MSD) and Free Music Archive dataset (FMA) for our prediction. For predicting the year label of single datasets and the combined dataset, a simple MLP model with two hidden layers and one output neuron is used, where the first hidden layer is twice as large as the input layer and the second hidden layer is half as large as the input layer.

For federated learning, we adopt the SplitNN~\cite{gupta_distributed_2018} model, which concatenates one layer of two separate models and passes it as the input to the third model. To make the federated method comparable to the non-federated one, we use the same MLP structure for each party when training for single datasets, and concatenate the second hidden layer of the two networks and connect the concatenated layer to the output neuron for prediction.

\paragraph{Recommendation}\label{par:recommendation}
Recommendation system has become a core component in various industrial applications such as product promotion and advertisement display{}. As the three groups of movie datasets listed in Table~\ref{tab:tasks} contain user information and movie information, we perform federated recommendations based on these datasets.

Taking the movies dataset as an example, we can align the MovieLens 1M~\cite{harper_movielens_2016} and IMDB Movie~\cite{maas-EtAl:2011:ACL-HLT2011} to augment the movie information in MovieLens 1M. We split the MovieLens 1M dataset into two parties, with one of them having the rating matrix and the other containing the auxiliary information of users and movies. The two parties can then be vertically federated to predict a user's preference for a movie. We use a variation of Neural Collaborative Filtering (NCF)~\cite{he_neural_2017} as our underlying model. 

\subsection{Reference Implementations}\label{sub:reference_implementations}

This paper benchmarks the performances of both horizontal and vertical FL systems. The reference implementations serve three-fold purposes. First, as infrastructures of FL, their performances reveal the capability of federated learning, especially when compared with non-federated algorithms. The comparison between different FL algorithms helps us understand the characteristics of each algorithm. Second, the analysis of those systems serves as a baseline to more complex ones derived from them, including the ones that incorporate additional mechanisms such as privacy or encryption. Last, the reference implementations act as a general framework, regulating the implementation and analysis of future algorithms.

Following the simplicity principle, we first set a baseline, and develop variants from the baseline according to the basic aspects of FL systems. The variants are made directly comparable to the baseline by only changing configurations of a single module. A hybrid reference implementation that changes multiple configurations simultaneously is presented at last to demonstrate how our benchmark can be used to analyze complex systems.

\subsubsection{Baseline}\label{ssub:baseline}

Reference implementations for baseline set cornerstones for subsequent experiments and analysis. For horizontal tasks, we use the FedSGD~\cite{pmlr-v54-mcmahan17a} algorithm as the baseline, with no additional modifications such as privacy and security. As the algorithm itself requires the participation of every party in all communication rounds, it minimizes the randomness introduced by the algorithm, making the baseline and subsequent comparison to baseline more reliable. For synthetic datasets, we use i.i.d splitting to ensure each party has the same amount of data and identical distributions. We consolidate parameters found by simple preliminary experiments into reference implementations so that the baseline is easily reproducible.

Additionally, we train datasets of each party separately and also add the ``combined'' experiment into the baseline, where the tested datasets are directly combined and treated as if they were a single dataset, and trained on the same model as the FL experiment. Through comparing these experiments, we can also explore some basic properties of the specific tasks, such as the improvement brought by the ``federation'' of federated learning.

For vertical tasks, We directly train a model on the combined dataset as the baseline, where datasets from different parties are directly aligned and treated as a single dataset. Due to the heterogeneity of different parties in vertical FL, we cannot directly use one parties' model or combine them, so we set the baseline model to resemble the target model in structure.

\subsubsection{Data Partitioning and Data Distribution}\label{ssub:non_i_i_d_ness}

As non-i.i.d-ness can greatly affect the quality of the model, we try to quantify the relation between them.

We focus on the relations between model accuracy and two types of non-i.i.d-ness: label distribution skew and quantity skew~\cite{kairouz_advances_2021}. These types of skewed datasets can be obtained by artificial partitioning and most empirical works focus on them~\cite{pmlr-v54-mcmahan17a}. The object classification task is used since manipulating the distribution of synthetic datasets is more simple and provides clearer results when compared to the baseline.

For both types of non-i.i.d-ness, we use Dirichlet distribution for partitioning, which has been widely used in the previous studies~\cite{yurochkin_bayesian_2019,Wang2020Federated,wang_tackling_2020}. In the label distribution skew scenario, we sample \(\mathbf{p}_{k}\sim Dir_{N}(\alpha)\) for each class \(k\) in the dataset and allocate \(p_{k,i}\) proportion of data records of class \(k\) to client \(i\). In the quantity distribution skew scenario, we sample \(\mathbf{p}\sim Dir_{N}(\alpha)\) and allocate \(p_{i}\) proportion of data records to client \(i\) with identical class distributions. In both scenarios, \(N\) denotes the number of clients and \(Dir_{N}(\alpha)\) is the \(N\)-dimensional Dirichlet distribution with concentration parameter of \(\alpha\). Smaller \(\alpha\) means more skewed distribution, potentially leading to worse results. Although we also provided functionalities to split the dataset by power law in the quantity skew scenario, we use Dirichlet distribution in our reference implementation for uniformity.

\subsubsection{FL Algorithms}\label{ssub:fl_fl_algorithms}

Comparing different FL algorithms gives us an idea of how tweaking the algorithms itself can affect the final model performance. For horizontal tasks, we compare four different algorithms (including the baseline): FedSGD~\cite{pmlr-v54-mcmahan17a}, FedAvg~\cite{pmlr-v54-mcmahan17a}, FedProx~\cite{li_federated_2020-1} and FedNova~\cite{wang_tackling_2020}, where for FedAvg, FedProx and FedNova we sample 50\% of the clients in each communication round. Compared to FedSGD, FedAvg added client sampling mechanisms, where only a fraction of clients are selected to compute the update in each communication round, FedProx introduced an additional \(\ell_{2}\) regularization term in the local objective to limit the distance between the local and global models, and FedNova introduced scales each local updated by the number of their local steps to mitigate the heterogeneity between updates. For vertical tasks, we compared baseline with two types of SplitNN: MLP based (for the year prediction task) and NCF based (for the recommendation task), to explore the performance of the SplitNN structure.

\subsubsection{Encryption}\label{ssub:impact_of_encryption}

Multiple ways can be used to perform encryption in FL, as stated in Section~\ref{sec:backgrounds}. However, through preliminary experiments, we found that these libraries are not suitable for directly being applied to various FL algorithms. As \citet{yang_batchcrypt_2019,xu_hybridalpha_2019} and other researchers have presented, their time and memory overhead are unacceptable without non-trivial optimization, since the large-number required by representing HE ciphertexts and computation complexity required by HE-based or garbled circuit-based SMC methods. So instead, we applied a simple scheme: SMC via trivial secret sharing, which is also used in some FL works regarding secure aggregation~\cite{bonawitz_practical_2016,bonawitz_practical_2017}.

Using this method, in a \(N\)-client setup, the gradient \(\mathbf{g}_{i}\) of client \(i\) are divided into \(K(2\leq K\leq N)\) parts:

\begin{equation}
    \mathbf{g}_{i}=\mathbf{g}_{i1}+\mathbf{g}_{i2}+\dots+\mathbf{g}_{iK}
\end{equation}

Where the first \(K-1\) parts are drawn from a uniform distribution by \(\mathbf{g}_{ij}\sim\mathcal{U}(0, Q)\) and \(Q\) is a large number. The \(j\)-th part \(\mathbf{g}_{ij}\) in these \(K-1\) splits are sent to client (\((i+j)\!\mod N\)) respectively via secure channels. The last part \(\mathbf{g}_{iK}\) is calculated by \(\mathbf{g}_{iK}=((Q+\mathbf{g}_{i}-\sum_{j=1}^{K-1}\mathbf{g}_{ij})\!\mod Q)\) and remains on client \(i\). After receiving each others' random split, each client \(i\) calculate the sum of gradient splits:

\begin{equation}
    \mathbf{g}_{i}'=\mathbf{g}_{iK}+\sum_{j=1}^{K-1}\mathbf{g}_{((K+i-j)\!\mod K)j}
\end{equation}

\(\mathbf{g}_{i}'\) is then sent to server for aggregation, as \(\sum_{i} \mathbf{g}_{i}'=\sum_{i} \mathbf{g}_{i}\). This method requires additional communication rounds but does not incur overwhelmingly high computational and memory cost, which is practical to implement in FL systems.

\subsubsection{Differential Privacy}\label{ssub:privacy}

The evaluation of differential privacy helps us understand how noise affects different types of workloads, and the intensity of their effects in FL settings compared to non-FL settings. In this reference implementation, we adopt the widely-used moments-accountant approach~\cite{abadi_deep_2016}, controlling the noise and privacy level with the \((\varepsilon,\delta)\) parameter. We focus on the final model accuracy, and convergence speed in this implementation, as they are the target-metrics that are most likely affected by DP noise.

\subsubsection{Communication Cost}\label{ssub:communication_cost}

We benchmark four types of compression and their impact on both the communication cost and model accuracy. To cover different types of methods, we implement a high-level compression method called adjusted local epochs, where more than one local epoch is performed in each communication round. We also implement 3 low-level compression methods, namely TopK, LowRank~\cite{vogels_powersgd_2019} and RandK. TopK only selects and sends \(K\) elements from the gradient with the largest absolute values, while LowRank utilizes matrix decomposition to reduce the size of the gradients. RandK selects and sends\(K\) elements from the gradient randomly, regardless of their distributions.

Similar to the differential privacy implementation, model accuracy and convergence speed are our major focuses. We also measure the uplink compression ratio in this implementation. By preliminary experiments, we found that it is hard to achieve a decent compression ratio by compressing the model in the broadcasting stage, so the downlink compression ratio is not our primary focus.

\subsubsection{Hybrid Examples}\label{ssub:hybrid_examples}

As real-world FL applications often incorporate multiple techniques, we set up an implementation in our benchmark to simulate such a scenario. We use both DP and SMC via secret sharing to enhance the security of the training process, and use both high-level and low-level compression methods to reduce communication costs. In addition to metrics used in the baseline, we also measure the time decomposition of those implementations to analyze the characteristics of these two scenarios and give suggestions on possible improvements of FL systems under that scenario.

\subsection{Applications}\label{sub:applications}

Our benchmark potentially has wide applications in the real world. The first potential application is to help developers to select the right algorithms for their applications. For example, if the goal of a developer is to utilize multiple companies' image data to efficiently train a classifier, the reference implementation associated results in our benchmark can help predict the performance of the implementation using existing statistics. The second potential application is to offer benchmark datasets and facilities for comparing different algorithms in a fair and systematic manner. If a researcher needs to implement a new FL algorithm, or wants to compare two existing algorithms whose original implementations require different experimental setups, our benchmark can be used as a framework to port those algorithms and provide a uniform comparison environment. We elaborate two potential applications of our benchmark suite.

\subsubsection{Algorithm Selection}\label{ssub:algorithm_selection}

As FL algorithms and their variants can perform differently on the same dataset, and different variants introduce different amount of computation and communication cost, balancing between various metrics such as accuracy, overhead and throughput has been a challenging task. By using and tweaking the reference implementations in our benchmark, these trade-offs can be measured and studied in a simple and reproducible way, thus helping the user select the algorithm according to their requirements.

\subsubsection{Algorithm Comparison}\label{ssub:algorithm_comparison}

Fairly comparing different algorithms or variants of different algorithms has long been an issue in the machine learning area, due to all kinds of setup differences, including the hardware setup, data splitting methods, differences in algorithm or module implementations, etc. To resolve those issues, our benchmark parameterizes and logs all setups, making fair comparison between algorithms and reproduction of the results an easy task.

%% file: sections/5-experiments.tex
\section{Preliminary Experiments}\label{sec:experiment}

This section presents preliminary evaluations to our reference implementations, with training setups described in Section~\ref{sub:reference_implementations}. The evaluations are divided into two parts. The horizontal FL experiments demonstrate all the reference implementations shown in Figure~\ref{fig:architecture}, including baseline (and improvement brought by federation), non-i.i.d-ness evaluation, algorithm comparison, impact of encryption, and differential privacy, and compression methods. The vertical FL experiments mainly focus on the performance improvement in the federated settings and the impact of the homomorphic encryption method.

To make the paper concise, only the baseline results of two workloads for each partitioning scheme are presented. To cover various types of tasks and models, we choose the object classification task and the sentiment analysis task for the horizontal FL experiments, while for vertical FL experiments we choose the year prediction task and the recommendation task with the movies dataset. Additional results for these tasks can be found in the appendix.

All the tasks use 83.33\% of the dataset for training, 8.33\% for testing, and 8.33\% for validation. The baseline experiments are repeated 12 times and uses Bayesian correlated t-test~\cite{corani_bayesian_2015,nilsson_performance_2018} to improve reproducibility of experiments and validity of results. For other experiments, each is repeated 3 times and both the mean and standard deviation for target metrics are reported.

The experiments are conducted on two Linux machines, each with a Xeon E5--2640 CPU @ 2.4GHz, 256GB DRAM, and three Nvidia GeForce RTX 2080 Ti GPU\@. Depending on the scale of the experiments, not all machines and GPUs are fully utilized. Only the minimum resources required are used to run each experiment.

\subsection{Baseline}\label{sub:baseline}

\paragraph{Experimental Setup}
Some important hyper-parameters not covered in Section~\ref{sub:reference_implementations} that greatly affect the experiment results or will be changed in subsequent reference implementations are presented in Table~\ref{tab:baseline_experimental_setups}. Trivial hyper-parameters that are not in the table are kept the same across tasks (including the not presented ones). To reduce redundancy, we only show the essential and altered ones in subsequent experiments, while omitting those that are kept the same as the baseline.

\begin{table}[htb]
    \centering
    \caption{Baseline experimental setups}\label{tab:baseline_experimental_setups}
    \begin{tabular}{l c c}
        \toprule
        Parameters                                    & Object Classification                                     & Sentiment Analysis \\
        \cmidrule(lr){1-3}
        \multirow{2}{*}{Underlying model}             & \multirow{2}{*}{\mrow{ResNet-18 with batch- \\normalization disabled}} & \multirow{2}{*}{\mrow{2-layer LSTM with\\sequence length of 200}}\\&&\\
        Data splitting                                & Synthetic, 5 parties, i.i.d                               & Realistic \\
        Accuracy metric                               & Top-1 accuracy                                            & Binary accuracy \\
        Optimizer                                     & SGD with momentum of 0.9                                  & Adam \\
        Learning rate scheduler                       & \multicolumn{2}{c}{Reduce on plateau, with factor of 0.1} \\
        Scheduler patience                            & 10                                                        & 20 \\
        Maximum communication rounds                  & \(\infty\)                                                & \(\infty\) \\
        Differential privacy \((\varepsilon,\delta)\) & \((\infty,\infty)\)                                       & \((\infty,\infty)\) \\
        Gradient norm clipping                        & \(\infty\)                                                & \(\infty\) \\
        Communication implementation                  & \multicolumn{2}{c}{Unix socket with no bandwidth limit} \\
        Communication compression method              & None                                                      & None \\
        \bottomrule
    \end{tabular}
\end{table}

\paragraph{Results and Analysis}

Each row in Table~\ref{tab:experiment-baseline} shows one of three types of training setups: training each dataset separately, training on the combined dataset from all parties, and using the FedSGD algorithm to train on all the datasets collaboratively. For non-synthetic datasets, accuracy is measured with the two types of test datasets: each party's own test dataset and the combined one created by concatenating all parties' test datasets (shown in the ``Combined'' column). The ``Combined'' column demonstrates a fair comparison between different training setups.

The baseline only focus on the ``FedSGD'' row and the ``Combined'' column, which compared with subsequent experiments and reference implementations. The other rows are columns are added to demonstrate the performance of the baseline itself.

\begin{table}[htb]
    \centering
    \caption{Baseline accuracy results with different test datasets}\label{tab:experiment-baseline}
    \addtolength{\tabcolsep}{-.3ex}
    \subtable[Object Classification]{
        \begin{tabular}{l c}
            \toprule
            \multirow{2}{*}[-.5ex]{\mrow[l]{Training \\ Setup}}
                     & Test Dataset \\ \cmidrule(lr){2-2}
                     & Combined \\ \cmidrule(lr){1-2}
            \multirow{2}{*}{\mrow[l]{Solo \\\hphantom{---} (1/5 datasets)}} & \multirow{2}{*}{78.54\%\(\pm\)0.81\%}         \\ & \\
            Combined & 88.92\%\(\pm\)0.76\% \\
            FedSGD   & \textbf{89.30\%\(\pm\)0.73\%} \\
            \bottomrule
        \end{tabular}
    }%
    \subtable[Sentiment Analysis]{
        \begin{tabular}{l c c c}
            \toprule
            \multirow{2}{*}[-.5ex]{\mrow[l]{Training \\Setup}}
                     & \multicolumn{3}{c}{Test Dataset} \\ \cmidrule(lr){2-4}
                     & IMDB                             & Amazon                        & Combined \\ \cmidrule(lr){1-4}
            IMDB     & 86.63\%\(\pm\)0.93\%             & 80.89\%\(\pm\)0.78\%          & 83.24\%\(\pm\)0.66\% \\
            Amazon   & 84.20\%\(\pm\)0.52\%             & 84.69\%\(\pm\)0.65\%          & 84.60\%\(\pm\)0.53\% \\
            Combined & \textbf{87.50\%\(\pm\)0.18\%}    & \textbf{86.13\%\(\pm\)0.47\%} & \textbf{86.81\%\(\pm\)0.26\%} \\
            FedSGD   & 87.07\%\(\pm\)0.92\%             & 85.62\%\(\pm\)1.21\%          & 86.34\%\(\pm\)1.03\% \\
            \bottomrule
        \end{tabular}
    }
    \subtable[Statistical Test Results]{
        \label{tab:experiment-baseline-test}
        \footnotesize
        \begin{tabular}{ll}
            \toprule
            Workload                             & Test Results \\
            \cmidrule(lr){1-2}
            \multirow{2}{*}{Obj. Classification} &
            \(p(\text{FedSGD},\text{Equal},\text{Solo})=\langle 1.00, 1.49\times 10^{-10}, 2.30\times 10^{-11}\rangle\), \\ &
            \(p(\text{FedSGD},\text{Equal},\text{Combined})=\langle7.97\times 10^{-2}, 9.17\times 10^{-1}, 3.29\times 10^{-3}\rangle\) \\
            \cmidrule(lr){1-2}
            \multirow{3}{*}{\mrow[l]{Sent. Analysis \\(with ``Combined'' test dataset)}} &
            \(p(\text{FedSGD},\text{Equal},\text{IMDB})=\langle 9.99\times10^{-1}, 9.42\times10^{-4}, 3.52\times10^{-6}\rangle\) \\ &
            \(p(\text{FedSGD},\text{Equal},\text{Amazon})=\langle 9.23\times 10^{-1}, 7.70\times10^{-2}, 7.26\times10^{-5}\rangle\) \\ &
            \(p(\text{FedSGD},\text{Equal},\text{Combined})=\langle 4.5\times10^{-3}, 8.54\times10^{-1}, 1.41\times10^{-1}\rangle\) \\
            \bottomrule
            \multicolumn{2}{p{\dimexpr0.83\textwidth}}{\(^{1}\) Here, \(p(\text{A},\text{Equal},\text{B})=\langle p_{\text{A}}, p_{\text{Equal}}, p_{\text{B}}\rangle\) means algorithm A being superior has the probability of \(p_{\text{A}}\), and algorithm B being superior has the probability of \(p_{\text{B}}\). The possibility of them being equal within \textit{rope} (\textit{rope}=\(\pm1\%\) in this table) is \(p_{\text{Equal}}\).}
        \end{tabular}
    }
\end{table}

In both tasks, our federated reference implementation outperforms training with each party's private data and is close to the training with the combined dataset. The difference between the FedSGD setup and the Combined setup is incurred by FL's transferring model parameters or gradients instead of training data, introducing a layer of abstraction and thus causing information loss. The result also shows that such information loss can be mitigated with proper hyperparameter tuning.

When comparing the results of FedSGD and Solo setup of the realistically portioned tasks such as sentiment analysis, we make two observations: 1) When the training and testing dataset comes from the same party (e.g. Training and testing are both using the IMDB dataset), the results describe the model quality. 2) When they come from different parties (e.g. Using the IMDB dataset for training and the Amazon dataset for testing), the results describe the generalization ability of the model. FedSGD improves both the quality of the models and the generalization ability of the models, achieving higher accuracy.

\begin{figure}[htpb]
    \centering
    \begin{minipage}[t]{0.36\textwidth}
        \centering
        \subfigure[FedSGD - Solo]{
            \includegraphics[width=0.5\linewidth]{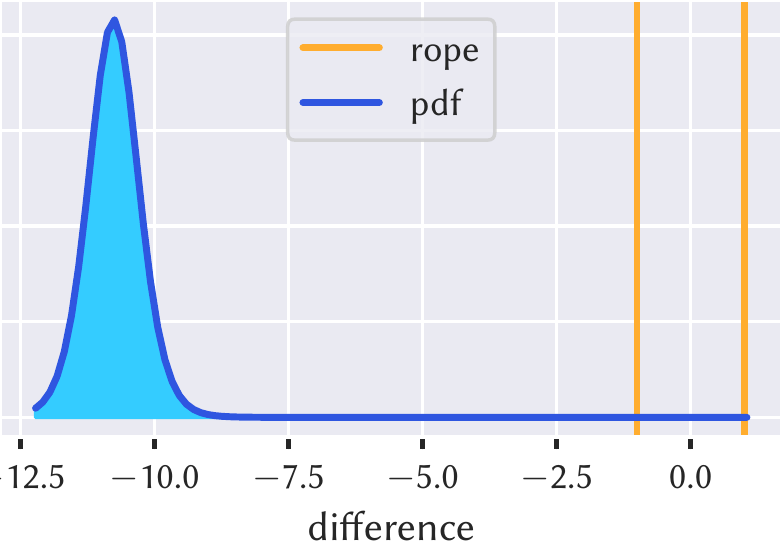}
        }%
        \subfigure[FedSGD - Combined]{
            \includegraphics[width=0.5\linewidth]{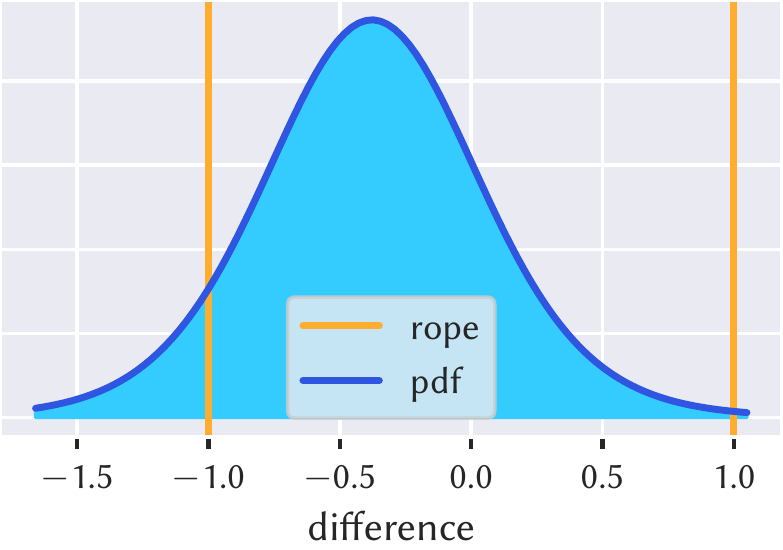}\label{fig:t-test-b}
        }
        \caption{Correlated t-test result of the object classification task}\label{fig:t-test-cifar}
    \end{minipage}
    \hfill
    \begin{minipage}[t]{0.54\textwidth}
        \centering
        \subfigure[FedSGD - IMDB]{
            \includegraphics[width=0.33\linewidth]{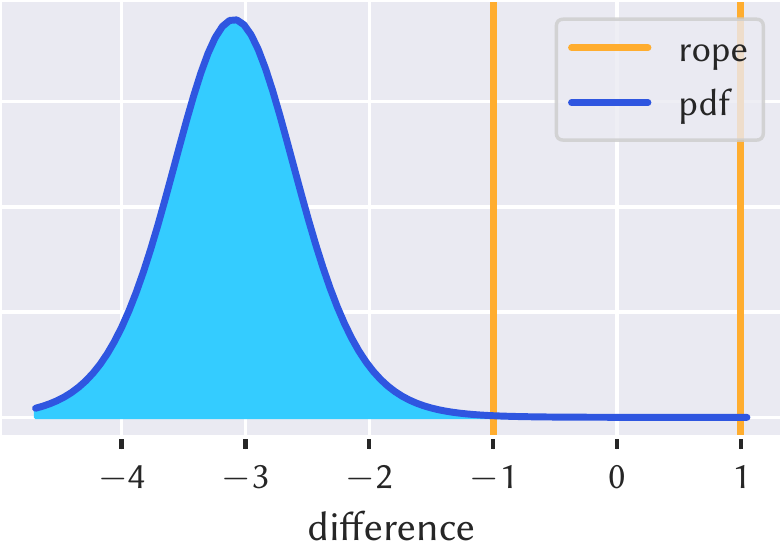}
        }%
        \subfigure[FedSGD - Amazon]{
            \includegraphics[width=0.33\linewidth]{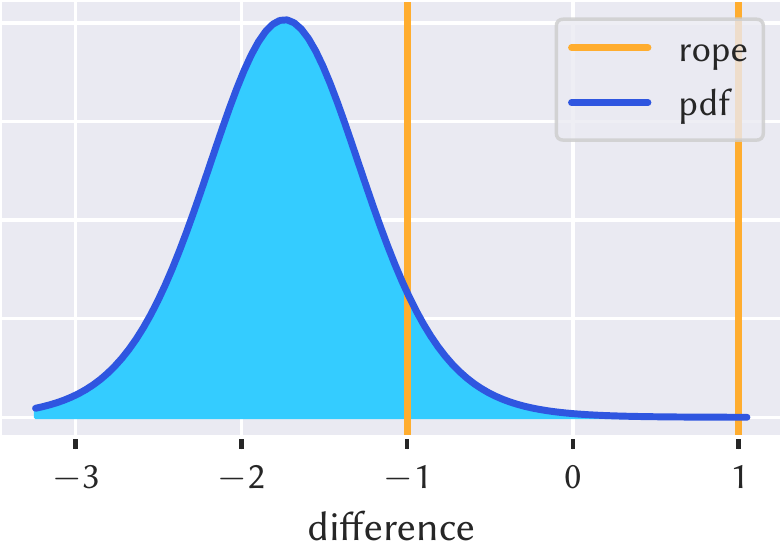}
        }%
        \subfigure[FedSGD - Combined]{
            \includegraphics[width=0.33\linewidth]{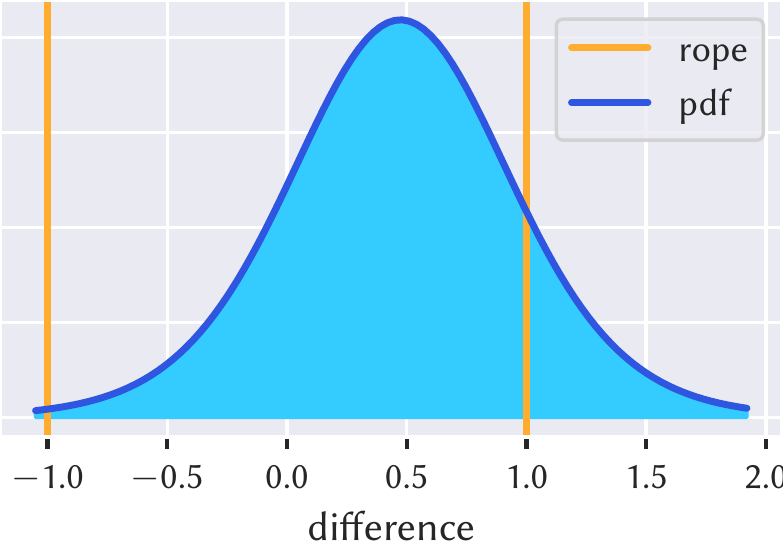}
        }
        \caption{Correlated t-test result of the sentiment analysis task}\label{fig:t-test-sent}
    \end{minipage}
    \hfill
\end{figure}

Figure~\ref{fig:t-test-cifar} and Figure~\ref{fig:t-test-sent} graphically explains the t-test results in Table~\ref{tab:experiment-baseline-test}. Each sub-figure corresponds to a row in Table~\ref{tab:experiment-baseline-test}, and presents the posterior probability density function of two algorithms' accuracy difference (derived by applying t-test to our experiment results). The area under the curve shows the exact probability of which algorithm is superior given a threshold, where the threshold is defined by a \textit{rope} value (shown as a pair of orange lines in the figure). Compared to directly testing a null-hypothesis, this test produces a distribution that can be queried to produce the exact probability, instead of just one p value indicating which one is better, thus being more informative.

Taking Figure~\ref{fig:t-test-b} as an example, the result \(p(\text{FedSGD},\text{Equal},\text{Combined})=\langle 7.97\times 10^{-2}, 9.17\times 10^{-1}, 3.29\times 10^{-3}\rangle\) means that the probability of FedSGD and the Combined setup achieves same accuracy (within \(\pm 1\%\) range) is 91.70\%, while the probability of FedSGD being superior is 7.97\%, and the probability of combined setup being superior is 0.33\%, so we can safely conclude that FedSGD and combined setup have very similar accuracy performance. Note that the test results depends on the selection of the ropes. For the rest of the result, we report the probability under a fixed rope value.

\begin{table}[htpb]
    \centering
    \caption{Other target metrics of baseline implementations}
    \subtable[Other target metrics of baseline implementations]{
        \label{tab:other_target_metrics}
        \begin{tabular}{l c c c c}
            \toprule
            \multirow{2}{*}{Training Setup}  &
            \multirow{2}{*}{\mrow{Convergence \\Time (rounds)}} &
            \multirow{2}{*}{\mrow{Throughput \\(samples/sec)}}     &
            \multirow{2}{*}{\mrow{Framework \\Overhead}} \\ &&& \\
            \cmidrule(lr){1-4}
            Object classification (FedSGD)   & 159.42\(\pm\)17.45 & 3913.31\(\pm\)36.85 & 25.83\%\(\pm\)0.61\% \\
            Object classification (Combined) & 85.58\(\pm\)5.04   & 1644.52\(\pm\)90.40 & 6.40\%\(\pm\)0.33\% \\
            \cmidrule(lr){1-4}
            Sentiment analysis (FedSGD)      & 176.67\(\pm\)53.25 & 1212.97\(\pm\)17.63 & 36.41\%\(\pm\)0.70\% \\
            Sentiment analysis (Combined)    & 127.67\(\pm\)28.89 & 860.54\(\pm\)41.98  & 16.50\%\(\pm\)0.84\% \\
            \bottomrule
        \end{tabular}
    }
    \subtable[Statistical Test Results]{
        \footnotesize
        \begin{tabular}{lll}
            \toprule
            Workload & Metric             & Test Results\(^{1}\) \\
            \cmidrule(lr){1-3}
            \multirow{3}{*}{Obj. Classification}
                     & Convergence Time   & \(p(\text{FedSGD},\text{Equal},\text{Combined})=\langle 1.00, 2.22\times 10^{-6}, 1.61\times 10^{-7}\rangle\) \\
                     & Throughput         & \(p(\text{FedSGD},\text{Equal},\text{Combined})=\langle 1.00, 1.28\times 10^{-14}, 7.88\times 10^{-15}\rangle\) \\
                     & Framework Overhead & \(p(\text{FedSGD},\text{Equal},\text{Combined})=\langle 1.00, 2.05\times 10^{-12}, 0.00\rangle\) \\
            \cmidrule(lr){1-3}
            \multirow{3}{*}{Sent. Analysis}
                     & Convergence Time   & \(p(\text{FedSGD},\text{Equal},\text{Combined})=\langle 9.13\times 10^{-1}, 6.15\times 10^{-2}, 2.46\times 10^{-2}\rangle\) \\
                     & Throughput         & \(p(\text{FedSGD},\text{Equal},\text{Combined})=\langle 1.00, 1.71\times 10^{-7}, 3.80\times 10^{-10}\rangle\) \\
                     & Framework Overhead & \(p(\text{FedSGD},\text{Equal},\text{Combined})=\langle 1.00, 9.51\times 10^{-11}, 5.55\times 10^{-16}\rangle\) \\
            \bottomrule
            \multicolumn{3}{l}{\(^{1}\) For convergence time, rope=\(\pm\)10. For throughput, rope=\(\pm\)100. For framework overhead, rope=\(\pm\)10\%.}
        \end{tabular}
    }
\end{table}

Table~\ref{tab:other_target_metrics} shows other target metrics of the baseline implementations. It is worth noting that the training stops when the learning rate is reduced the 4th time, and the convergence time is measured by the number of communication rounds under this assumption. The throughput is the total end-to-end throughput of all clients, and the framework overhead measures all the time not spent on training and testing the model, which is calculated by  \((t_{\text{total}}-t_{\text{train}})/t_{\text{total}}\), where \(t_{\text{total}}\) is the end-to-end program execution time, and \(t_{\text{train}}\) includes time spent on training and testing.

In the results, compared to the ``Combined'' setup, FedSGD requires more rounds to converge and incurs more time overhead, but it benefits from the high parallelism and achieves higher end-to-end throughput. As the bandwidth in the baseline is not limited, the communication overhead only occupies a trivial portion of the framework overhead. This indicates that solving the asynchronous problem can be a effective way to improve the performance of FL\@. Detailed time decomposition in more realistic scenarios can be found in Section~\ref{sub:hybrid_example}.

\textit{Summary:} Improvements can be expected regarding both accuracy of and generalization ability of the model. The accuracy of the FL model is close or even better than the model trained with the combined dataset. Horizontal FL algorithms also achieve higher throughput compared to the solo setup, which can be used to boost the performance of ML training if the convergence rounds can be further reduced.

%% file: sections/6-conclusions.tex
\section{Conclusions and Future Works}\label{sec:conclusion}

In this paper, we propose the OARF framework, which serves as a benchmark suite and also a framework for FL\@. We collect datasets from different sources to assemble realistic FL tasks and use them in our reference implementations to quantitatively reveal interesting properties of FL systems. Most importantly, we provide generalized methods and metrics, and datasets from different sources to benchmark and model complex FL systems. Using these methods, our system can easily be extended to suite future research and development demands. Through the evaluation of the reference implementations, we demonstrated its capability and opportunities as follows:

\begin{itemize}
    \item The layered modular design allows substitution of datasets, models and various add-ons at different levels, making the benchmark suit versatile for different application requirements.
    \item The benchmark system includes metrics to comprehensively cover different aspect of the workloads, and can be substituted or extended easily to support new demands.
    \item The reference implementation provided by the benchmark are effective enough to serve as performance references for FL applications.
\end{itemize}

Despite the capabilities and opportunities, our benchmark also shows some limitations. First, the supported algorithms can be greatly extended. More complex algorithms such as FedMA~\cite{Wang2020Federated} can be added to the implementation. We are working on this, and meanwhile studying collaborative mechanisms to attract other developers to contribute their algorithms to this benchmark. The second is that modules of the benchmark can be expanded and optimized. For example, better alignment techniques for aligning the vertical datasets can be implemented, and optimizations to improve the encryption speed can be applied or added as an option. Third, we are developing utilities to help interpret the results can be integrated into the benchmark suite. Fourth, the benchmark workload needs to evolve as the development of federated learning.

The reference implementations in the benchmark suite are by no means complete. In addition to making this benchmark suite open to the community, we also plan to add more modules and reference implementations in the future. Adapting and generalizing more algorithms to make them work with existing modules is our future work. Hybrid FL is also an area worth investigating.

%% file: sections/11-appendix.tex
\section{Differential Privacy}\label{ssub:differential_privacy}

While SMC and HE protect against information leakage, they cannot prevent inference attack~\cite{shokri_membership_2017}, especially those that use additional data, where the adversary tries to recover information such as user identities from models after they are finally released. And this is where \emph{differential privacy} (DP)~\cite{hutchison_calibrating_2006} becomes important. It protects individuals' privacy in the released model by ensuring the presence or not of each individuals' record will not affect the output of the model much. Taking \((\varepsilon,\delta)\)-DP as an example, given two datasets \(D, D'\in\mathcal{D}\) that only differs in one entry, and a mechanism \(\mathcal{M} : \mathcal{D}\to\mathbb{R}\) that operates on the dataset, for any set \(S\subseteq\mathbb{R}\), \((\varepsilon,\delta)\)-DP is formulated by inequality:

\begin{equation}
    \Pr[\mathcal{M}(D)\in S] \leq e^{\epsilon}\cdot\Pr[\mathcal{M}(D')\in S] + \delta
\end{equation}

The inequality states that given a mechanism the difference of the probability that the result is in set \(S\) is bounded by a variables \(\varepsilon\) and \(\delta\), where \(\delta\) is zero for \(\epsilon\)-differential privacy and a non-zero value for \((\epsilon,\delta)\)-differential privacy. When \(\varepsilon\) And \(\delta\) is sufficiently small, one can hardly tell from the result of operation \(\mathcal{M}\) whether the entry that differs in \(D\) and \(D'\) is inside the database, thus guaranteeing individuals' privacy.

Such guarantee is often achieved by adding random noises to the data or the model. Gaussian noise is often used due to its advantage over other types of noises in the statistical analysis of its impact on the mechanism~\cite{dwork_algorithmic_2013}. Thus, through the two parameters \(\varepsilon\) and \(\delta\), we can quantify the privacy loss and the amount of noise that needs to be added. Unlike SMC and HE, DP often introduces marginal computational overhead but has a negative impact on the accuracy of the model, due to the noise being added. As the relation between DP parameters and model accuracy has only been systematically tested in the machine learning setting~\cite{abadi_deep_2016}, benchmarking DP in the FL setting will set a good baseline and provide useful insights on reducing its negative impact on model accuracy.

\section{Preliminary Experiments}\label{sec:appendix_experiments}

This section completes the preliminary experiments of the reference implementation for Section~\ref{sec:experiment}. All horizontal FL experiments are conducted in the same environment as the baseline experiment, and the setups are identical with the baseline as shown in Table~\ref{tab:baseline_experimental_setups} if not explicitly stated.

\subsection{Data Partitioning and Data Distribution}\label{sub:non_i_i_d_ness}

\paragraph{Experimental Setup}
As stated in Section~\ref{sub:reference_implementations}, we change the data splitting method of the object classification task with two types of methods: label distribution skew and quantity skew. The experimented \(\alpha\) value of the Dirichlet distribution in each method ranges from 0.2 to 1.0, where smaller \(\alpha\) value stands for higher non-i.i.d-ness.

\paragraph{Results and Analysis}

\begin{figure}[htb]
    \centering
    \subfigure[Model accuracy]{
        \includegraphics[height=6.4\baselineskip]{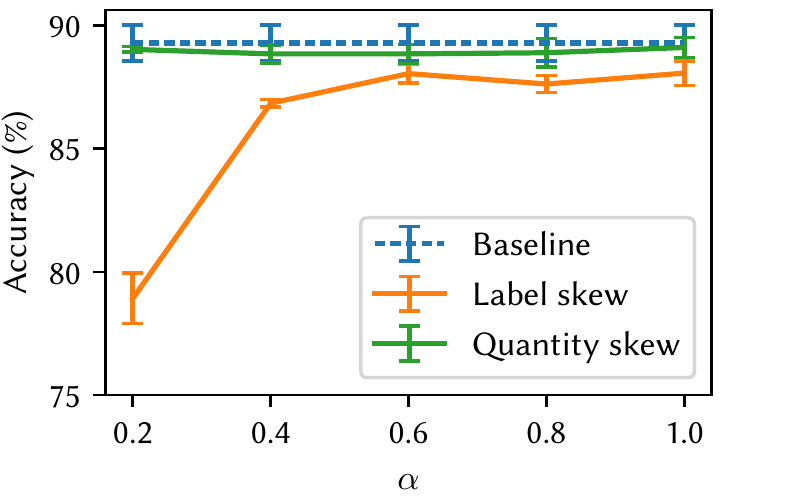}\label{fig:experiment-noniid-accuracy}
    }%
    \subfigure[Throughput]{
        \includegraphics[height=6.4\baselineskip]{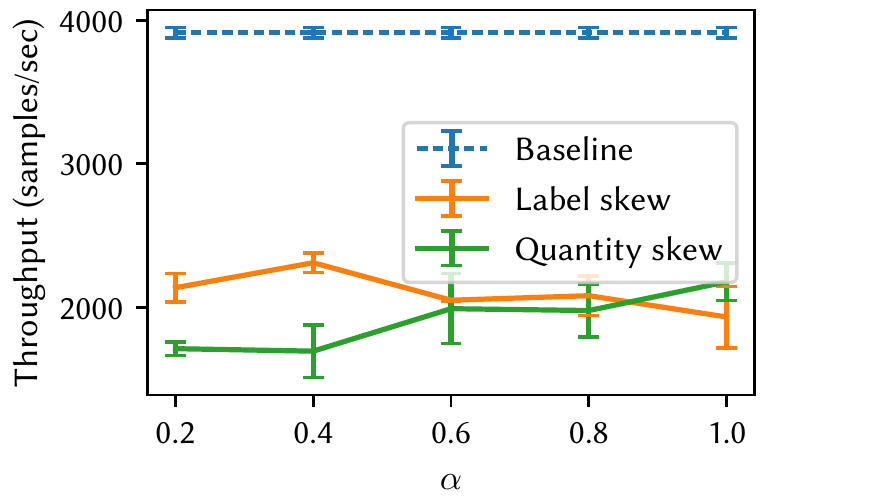}\label{fig:experiment-noniid-throughput}
    }%
    \subfigure[Framework time overhead]{
        \includegraphics[height=6.4\baselineskip]{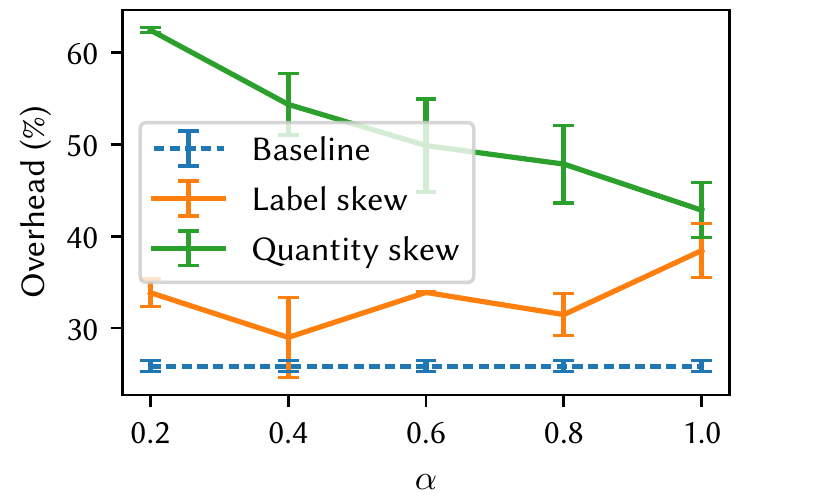}\label{fig:experiment-noniid-overhead}
    }
    \caption{Results of non-i.i.d experiments}\label{fig:experiment-noniid}
\end{figure}

Figure~\ref{fig:experiment-noniid-accuracy} shows the accuracy result of these experiments. Both label skew and quantity skew harms the model accuracy compared to be baseline result. However, quantity skew has less impact compared to label skew and is also less sensitive to the skewness. This result is consistent with other studies~\cite{li_federated_2021}, which indicates that future research works can put their focus on various label skew scenarios. Another interesting result shown in Figure~\ref{fig:experiment-noniid-overhead} is that quantity skew has more impact on framework overhead. This is within our expectation since FedSGD is a synchronized program, and more quantity skewness means more time spent on waiting for all clients to synchronize, resulting in significant overhead. From Figure~\ref{fig:experiment-noniid-throughput} it can also be observed that both types of the non-i.i.d-ness harm throughput, but there is no strong correlation between them.

\textit{Summary}: Various types of Non-i.i.d-ness play have a significant impact on the model accuracy. In-depth analysis has been made toward this issue both theoretically~\cite{zhao_federated_2018} and empirically~\cite{li_federated_2021}. But an effective method to mitigate it, especially under the FL scenario, still needs exploration.

\subsection{Federated Learning Algorithms}\label{sub:fl_algorithms}

\paragraph{Experimental Setup}
This implementation compares different FL algorithms, including FedSGD (baseline), FedAvg, FedNova, and FedProx. The client sampling ratios of each communication round are set to 0.4 (2 parties) and 0.2 (1 party) in the object classification task and sentiment analysis task respectively when using the FedAvg, FedNova, and FedProx algorithm. As FedProx introduces a new hyper-parameter \(\mu\) which controls the distance between each parties' local model and the global model, we experiment with two different values \(\mu=0.01\) and \(\mu=0.001\) and compare their results.

\paragraph{Results and Analysis}

Table~\ref{tab:experiment-algorithms} shows the accuracy results of different FL algorithms. Compared to the baseline, other algorithms do not achieve higher accuracy as not all clients are involved in gradient computation in each communication round. Compared to FedAvg, both FedNova and FedProx achieve slightly higher accuracy on the object classification task, but their performance is almost indistinguishable performance on the sentiment analysis task.

\begin{table}[htb]
    \centering
    \caption{Accuracy results with different FL algorithms}\label{tab:experiment-algorithms}
    \addtolength{\tabcolsep}{-.2ex}
    \subtable[Object Classification]{
        \begin{tabular}{l c c}
            \toprule
            Algorithm     & Accuracy                      & Throughput \\
            \cmidrule(lr){1-3}
            FedSGD        & 89.30\%\(\pm\)0.73\%          & \textbf{3913.31\(\pm\)36.85} \\
            FedAvg        & 88.74\%\(\pm\)0.43\%          & 1757.60\(\pm\)42.56 \\
            FedNova       & 88.78\%\(\pm\)0.26\%          & 1742.07\(\pm\)97.50 \\
            FedProx\(^1\) & \textbf{89.36\%\(\pm\)0.08\%} & 1500.51\(\pm\)63.24 \\
            FedProx\(^2\) & 88.96\%\(\pm\)0.41\%          & 1536.98\(\pm\)11.87 \\
            \bottomrule
        \end{tabular}
    }%
    \subtable[Sentiment Analysis]{
        \begin{tabular}{l c c}
            \toprule
            Algorithm     & Accuracy                      & Throughput \\
            \cmidrule(lr){1-3}
            FedSGD        & 86.34\%\(\pm\)1.03\%          & \textbf{1212.97\(\pm\)17.63} \\
            FedAvg        & 86.64\%\(\pm\)0.03\%          & 778.68\(\pm\)2.52 \\
            FedNova       & 86.21\%\(\pm\)0.11\%          & 802.03\(\pm\)30.54 \\
            FedProx\(^1\) & \textbf{86.74\%\(\pm\)0.07\%} & 760.57\(\pm\)1.28 \\
            FedProx\(^2\) & 86.47\%\(\pm\)0.19\%          & 777.51\(\pm\)26.72 \\
            \bottomrule
        \end{tabular}
    }
    \footnotesize
    \flushleft
    \par \(^{1}\mu=0.01\), \(^{2}\mu=0.001\)
\end{table}

As for the throughput, all the tested algorithm has lower throughput than the baseline due to the client selection mechanism in these algorithms, and more time is spent on idling and waiting for synchronization. The framework overhead of FedAvg, FedNova, and FedProx does not vary much. For the object classification task, the average overhead for all the algorithms are 70.76\%\(\pm\)0.84\%, and for the sentiment analysis task, it is 63.12\%\(\pm\)0.61\%, both significantly higher than the baseline, which is also a result of extended idle time caused by client selection the mechanism and synchronization in these algorithms.

\textit{Summary:} The level of impact on the accuracy of various FL algorithms is task-specific. These algorithms perform similarly when applied to the tasks presented in this paper, but their performance is also dependent on other factors such as the quantity, quality and distribution of the datasets, and needs to be further compared in more scenarios.

\subsection{Encryption}\label{sub:secure_multiparty_computation}

\paragraph{Experimental Setup}\label{par:experimental_setup}
This experiment adopt SMC via trivial secret sharing as described in Section~\ref{sub:reference_implementations}. For the object classification task, as there are 5 parties, two different setups are tested where the number of splits of the gradient for secret sharing is 2 and 4 respectively. For the sentiment analysis task, The number of splits is only set to be 2 as there are only 2 parties involved in the training.

\paragraph{Results and Analysis}\label{par:results_and_analysis}

Table~\ref{tab:experiment-encryption} presents evaluation to the target metrics in the experiments. As the encryption method is lossless, the accuracy results and convergence time results stay in the same range compared to the baseline. The end-to-end throughputs are also not affected much when the number of splits is 2 and drop slightly when the number of splits is 4.

\begin{table}[htb]
    \centering
    \caption{Target metrics of the SMC reference implementation}\label{tab:experiment-encryption}
    \begin{tabular}{l c c c c}
        \toprule
        \multirow{2}{*}{Training Setup} &
        \multirow{2}{*}{\mrow{Convergence \\Time (rounds)}} &
        \multirow{2}{*}{\mrow{Throughput \\(samples/sec)}}     &
        \multirow{2}{*}{\mrow{Framework \\Overhead}} \\ &&& \\
        \cmidrule(lr){1-4}
        Object classification (split=4) & 187.00\(\pm\)2.83  & 1891.69\(\pm\)0.64 & 35.03\%\(\pm\)0.25\% \\
        Object classification (split=2) & 163.67\(\pm\)16.74 & 2021.11\(\pm\)5.11 & 29.68\%\(\pm\)0.13\% \\
        \cmidrule(lr){1-4}
        Sentiment analysis (split=2)    & 102.33\(\pm\)8.26  & 1011.00\(\pm\)9.93 & 26.37\%\(\pm\)0.23\% \\
        \bottomrule
    \end{tabular}
\end{table}

The framework overhead reflects the portion of extra time spent on encryption-related calculation and communication. Compared to the baseline, a non-negligible overhead is introduced by encryption and is proportional to the number of splits, since a larger number of splits causes longer time calculating each split and transferring them.

It is worth mentioning that SMC cannot effectively prevent involving parties knowing each other's gradient when using FedSGD/FedAvg-based algorithms in a two-party configuration, as one party can simply calculate the weight of the other party by subtracting the weight of the global model by its own weight. In that case, SMC can only protect against a malicious server instead of malicious clients.

\textit{Summary:} For horizontal FL, SMC via secret sharing is a practical method to prevent information leakage, with the cost of throughput decrease. But it can be challenging to apply it to vertical settings.

\subsection{Differential Privacy}\label{sub:differential_privacy}

\paragraph{Experimental Setup}
Compared to the baseline, we limit the maximum number of communication rounds to 300 and use a gradient clipping of 0.1 for both tasks according to preliminary experiments and analysis. The parameters are configured so the each party's training process satisfies \((\varepsilon,\delta)\)-differential privacy where \(\delta\) is fixed at \(\min(10^{-5}, 1/N)\), \(N\) being the number of samples in the dataset~\cite{dwork_algorithmic_2013} and vary \(\varepsilon\).

\paragraph{Results and Analysis}

\begin{figure}[htb]
    \centering
    \subfigure[Object Classification]{
        \includegraphics[height=8\baselineskip]{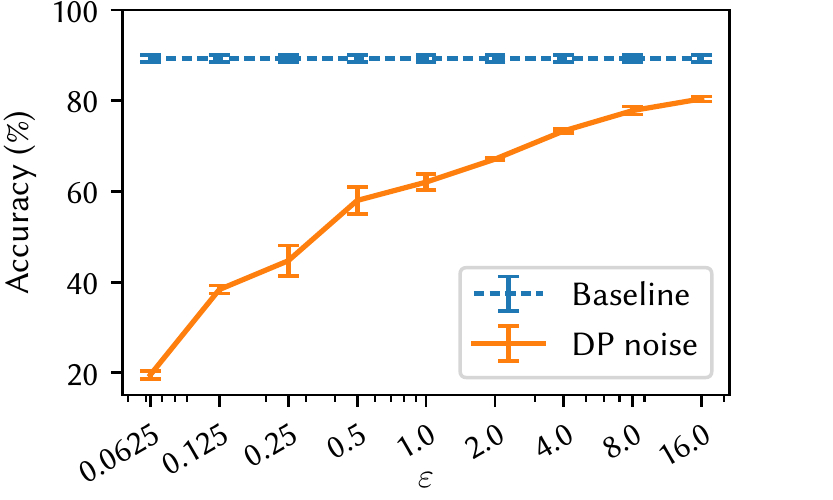}
    }%
    \subfigure[Sentiment Analysis]{
        \includegraphics[height=8\baselineskip]{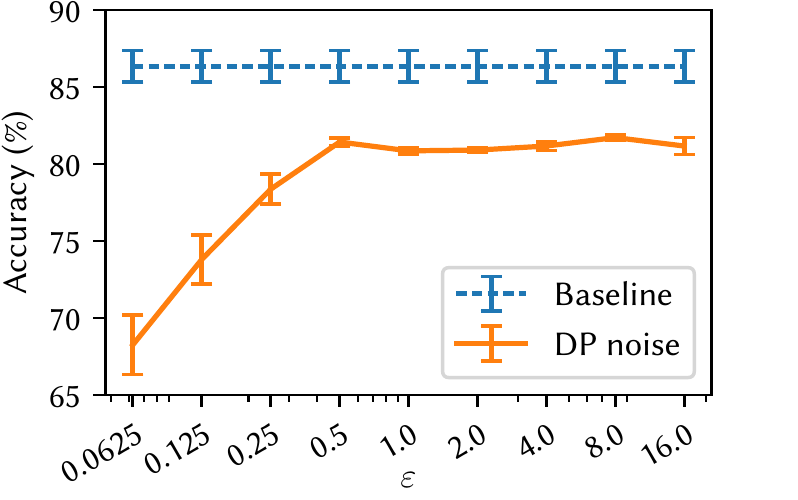}
    }
    \caption{Accuracy result under various levels of DP noise}\label{fig:experiment-dp-accuracy}
\end{figure}

As shown in Figure~\ref{fig:experiment-dp-accuracy}, the accuracy grows with the privacy budget \(\varepsilon\), which matches the intuition that the less noise added, the model becomes less private but its accuracy increases. Considering that an \(\varepsilon\) value larger than 2.0 may be unacceptable in practice~\cite{agrawal_differential_2008}, from the result, there is still much improvement space in terms of accuracy when using DP in FL settings.

Different types of tasks also have different sensitivities to DP noise. The accuracy of the object classification task dropped from 80.42\%\(\pm\)0.54\% to 58.00\%\(\pm\)2.99\% when \(\varepsilon\) dropped from 16.0 to 0.5, while the accuracy of the sentiment analysis tasks stays almost unchanged.

The throughput and framework overhead do not vary much when using different DP setups. For all 17 runs, the average throughput of object classification task and sentiment analysis task is 1041.07\(\pm\)3.73 samples per second and 1273.15\(\pm\)3.81 samples per second respectively, and the overheads are 11.61\%\(\pm\)0.08\% and 29.7\%\(\pm\)2.05\%. Compared to the baseline, the DP mechanism introduced a non-negligible amount of extra computing time. The convergence time varies greatly among different runs even with the same amount of DP noise, and there is no strong correlation between them.

\textit{Summary:} Differential privacy noise has a large negative impact on the accuracy of the model. The experiments show that there is still great potential to further improve it under practical privacy level, possibly by discovering composition methods with tighter privacy bounds~\cite{bun_concentrated_2016} or utilizing the internal noise~\cite{hyland_empirical_2020} generated in the training process.

\subsection{Communication Cost}\label{sub:communication_cost}

\paragraph{Experimental Setup}
Two types of methods are used for reducing the communication cost. For the high-level method, The number of local epochs is set to be 2, 4, 6, 8, and 10 for each communication round, and for the low-level compression methods, we adjust the settings for TopK and LowRank so that they achieve similar compression ratios. According to the difference between distributed learning and federated learning, we adjust the error feedback method for the LowRank algorithm from the work of~\citet{vogels_powersgd_2019}, multiplying the error by 0.5 before each feedback to achieve better convergence and accuracy.

\paragraph{Results and Analysis}

\begin{figure}[htb]
    \centering
    \subfigure[Object Classification]{
        \includegraphics[height=8\baselineskip]{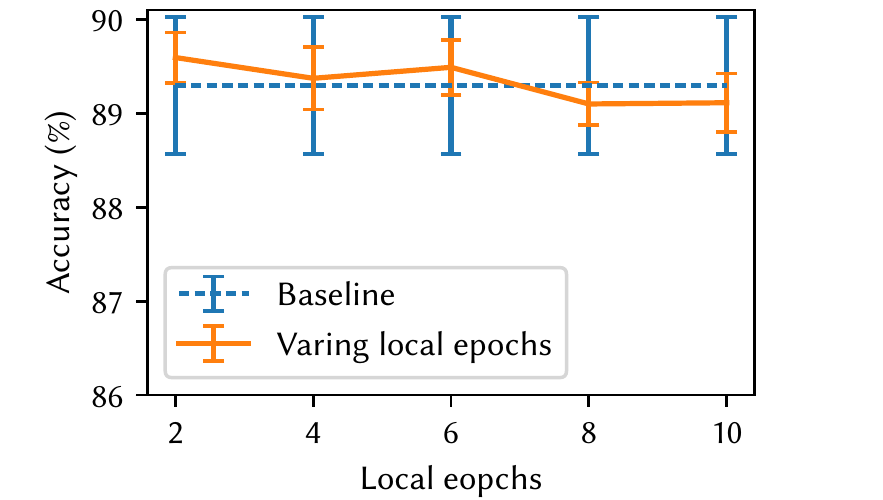}
    }%
    \subfigure[Sentiment Analysis]{
        \includegraphics[height=8\baselineskip]{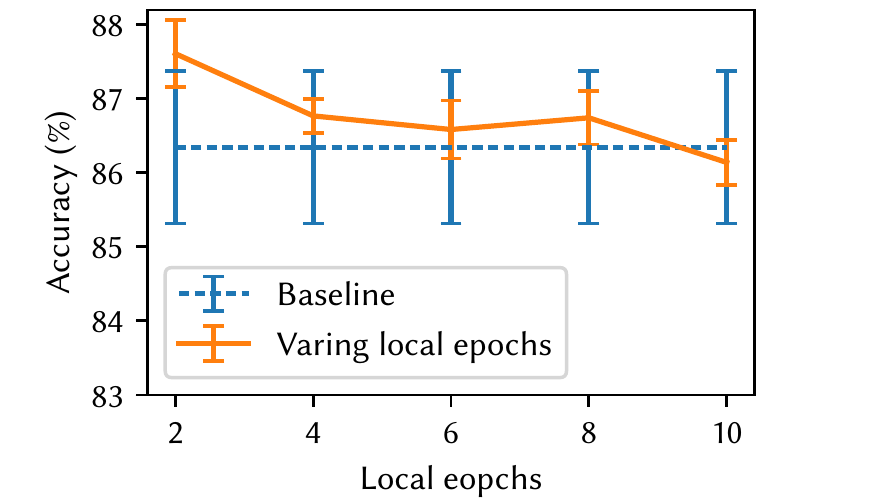}
    }
    \caption{Accuracy result with various number of local epochs}\label{fig:experiment-communication-high}
\end{figure}

Figure~\ref{fig:experiment-communication-high} shows the accuracy result of various local epochs per communication round. As the number of local epochs grows, the number accuracy drops slightly, but the drop is negligible for the object classification task and maintained within 2\% for the sentiment analysis task. However, given that the converges rounds are longer than the baseline on average, this simple method cannot effectively reduce the communication cost from an overall perspective. More fine-grained client selection-based methods that integrate into the algorithm such as FedAvg~\cite{pmlr-v54-mcmahan17a} or CMFL~\cite{wang_cmfl_2019} are essential if these types of high-level methods are desired.

The accuracy results of low-level methods in Table~\ref{tab:experiment-communication-low} show that these methods can effectively reduce the communication cost, with a cost of harming the accuracy. For the same method as TopK, the accuracy is strongly related to the sampling ratio, and for different compression methods, the accuracy can vary significantly.

\begin{table}[htb]
    \centering
    \caption{Accuracy results with different compression methods}\label{tab:experiment-communication-low}
    \addtolength{\tabcolsep}{-.2ex}
    \subtable[Object Classification]{
        \begin{tabular}{l c c}
            \toprule
            \multirow{2}{*}{\mrow[l]{Training \\Setup}} &
            \multirow{2}{*}{\mrow{Uplink \\ratio}} &
            \multirow{2}{*}{Accuracy} \\ &&\\
            \cmidrule(lr){1-3}
            Baseline       & 1.0    & 89.30\%\(\pm\)0.73\% \\
            TopK (k=1.0\%) & 33.32  & 84.78\%\(\pm\)0.08\% \\
            TopK (k=0.3\%) & 110.97 & 80.20\%\(\pm\)0.44\% \\
            LowRank (r=3)  & 93.98  & 78.35\%\(\pm\)0.18\% \\
            RandK (k=1\%)  & 99.89  & 39.92\%\(\pm\)1.41\% \\
            \bottomrule
        \end{tabular}
    }%
    \subtable[Sentiment Analysis]{
        \begin{tabular}{l c c}
            \toprule
            \multirow{2}{*}{\mrow[l]{Training \\Setup}} &
            \multirow{2}{*}{\mrow{Uplink \\ratio}} &
            \multirow{2}{*}{Accuracy} \\ &&\\
            \cmidrule(lr){1-3}
            Baseline       & 1.0    & 86.34\%\(\pm\)1.03\% \\
            TopK (k=1.0\%) & 33.29  & 75.84\%\(\pm \)2.16\% \\
            TopK (k=0.3\%) & 110.69 & 69.29\%\(\pm\)0.86\% \\
            LowRank (r=3)  & 45.34  & 84.40\%\(\pm\)0.07\% \\
            RandK (k=1\%)  & 99.69  & Cannot converge \\
            \bottomrule
        \end{tabular}
    }
\end{table}

Like high-level methods, the low-level methods also require more rounds to converge. However, the high compression ratio mitigates the overhead of extra communication rounds. For example, for the object classification task, TopK with \(k=1.0\%\) requires an average of 237.66\(\pm\)8.21 rounds to converge but achieves an uplink compression ratio of 33.32 each round, thus achieving an end-to-end uplink compression ratio of 25.42.

In this experiment, data is not compressed in the model broadcasting stage since the method of compressing models are very different from compressing gradient and often achieves worse performance according to our preliminary tests, so the downlink compression ratio of each communication round remains 1.0. In real-world applications, however, the downlink compression ratio is also an important factor to consider and significantly affects overall performance.

\textit{Summary:} The overall communication cost cannot be reduced by simply changing the number of local epochs in each communication round. For high-level methods, more fine-grained methods such as CMFL~\cite{wang_cmfl_2019} need to be considered when communication cost is critical. Low-level methods such as various types of gradient compression methods can effectively reduce per-epoch communication cost and overall communication cost, but the compression ratio and model accuracy need to be carefully balanced.

\subsection{Hybrid Example}\label{sub:hybrid_example}

\paragraph{Experimental Setup}
To analyze more complex setup and real-world scenarios, this experiment applied encryption (splits=2), DP (\(\varepsilon=1\)), TopK compression (k=1\%) to both tasks, and limited the socket bandwidth to 100Mbps to simulate a real-world scenario. We then analyze the time decomposition of the training process. The baseline is also re-run under the same bandwidth limit for a fair comparison.

\begin{figure}[htb]
    \centering
    \subfigure[Object Classification]{
        \includegraphics[height=8\baselineskip]{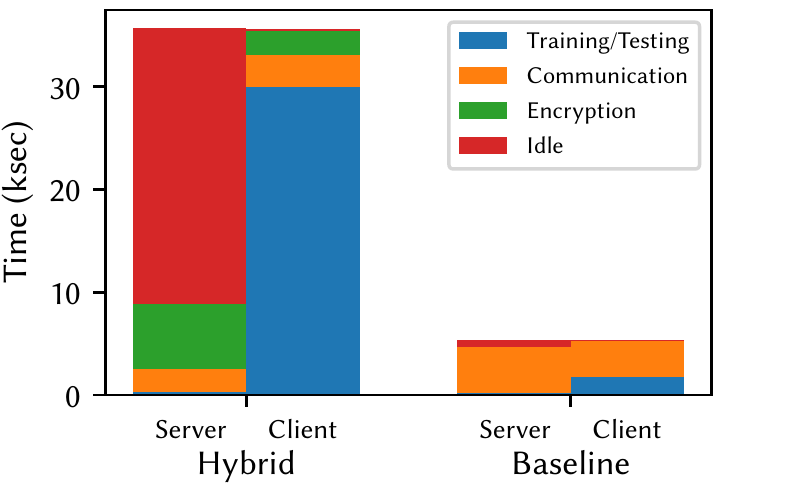}
    }%
    \subfigure[Sentiment Analysis]{
        \includegraphics[height=8\baselineskip]{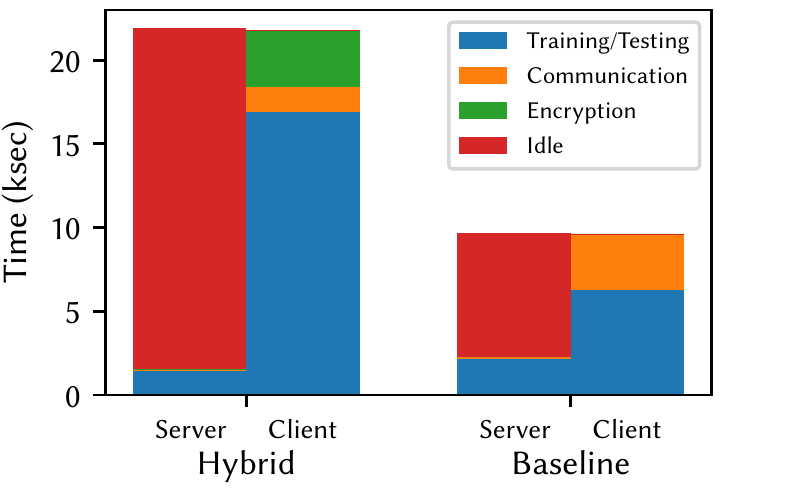}
    }
    \caption{Time decomposition of hybrid implementations}\label{fig:experiment-hybrid}
\end{figure}

\paragraph{Results and Analysis}
Figure~\ref{fig:experiment-hybrid} shows the time decomposition of two tasks. The ``Client'' bars in the chart show the average time decomposition for all involved clients. For both tasks, the idle time dominates the whole training process of the server, and training time dominates the training process of the clients. This means that synchronized algorithms like FedAvg cannot efficiently utilize the computation resource on the server.

Comparing the hybrid setup and the baseline, the most obvious difference is the prolonged training time, which is mostly contributed by the privacy mechanism and compression method. We also find that the overhead of encryption is non-negligible. Also, compression methods effectively reduce the communication cost.

Comparing the two tasks, it is obvious that in the object classification task, the server spends more time on encryption and communication compared to the sentiment analysis task. This signals that the difference in model size has a significant impact on the training process, as the model size of the object classification task is more than 15 times larger than the size of the sentiment analysis task (44.7MB compared to 2.9MB), the server requires more time to send and receive the message.

\textit{Summary:} The hybrid experiments suggest that idle time takes a significant portion in the centralized FL algorithms we have tested. Utilizing this part of idle computational power by offloading tasks to the server, or by developing asynchronous or even decentralized FL algorithms can be an interesting direction for future research.

\subsection{Vertical Federated Learning\@: Baseline}\label{sub:improvement_of_federation_in_vertical_federated_learning}

\paragraph{Experimental Setup}
This experiment test both the year prediction task and the recommendation task with the movies dataset. Table~\ref{tab:baseline-setup-vfl} shows the key configurations of the baseline. For the recommendation task, only the MovieLens dataset has both user data and movie data. As the IMDB dataset has no user data, we use it solely as auxiliary data in the combined setup and federated setup.

For the songs dataset, the samples in different datasets are linked with track names, and for the movies dataset, they are linked with movie titles. For those samples that have no counterpart in the other dataset, we pad the missing features with zeros.

\begin{table}[htb]
    \centering
    \caption{Vertical FL baseline experimental setups}\label{tab:baseline-setup-vfl}
    \begin{tabular}{l c c}
        \toprule
        Parameters              & Year Prediction                                           & Recommendation \\
        \cmidrule(lr){1-3}
        Single-party model      & 3-layer MLP                                               & DLRM~\cite{naumov_deep_2019} \\
        Accuracy metric         & Mean average error                                        & Mean squared error \\
        Optimizer               & Adam                                                      & Adam \\
        Learning rate scheduler & \multicolumn{2}{c}{Reduce on plateau, with factor of 0.1} \\
        Scheduler patience      & 10                                                        & 5 \\
        \bottomrule
    \end{tabular}
\end{table}

\paragraph{Results and Analysis}

\begin{table}[htb]
    \centering
    \caption{Baseline accuracy results}\label{tab:experiment-vert-baseline}
    \addtolength{\tabcolsep}{-.3ex}
    \subtable[Year Prediction]{
        \begin{tabular}{l c c}
            \toprule
            \mrow[l]{Training \\Setup}          &
            \mrow{Accuracy \\(MAE)}          &
            \mrow{Throughput \\(samples/sec)} \\
            \cmidrule(lr){1-3}
            FMA      & \textbf{3.158\(\pm\)0.123} & 26295.83\(\pm\)849.10 \\
            MSD      & 7.359\(\pm\)0.045          & \textbf{39333.34\(\pm\)980.12} \\
            Combined & 6.104\(\pm\)0.59           & 23111.17\(\pm\)963.63 \\
            SplitNN  & 6.083\(\pm\)0.135          & 19627.89\(\pm\)1131.67 \\
            \bottomrule
        \end{tabular}
    }%
    \subtable[Movie Recommendation]{
        \begin{tabular}{l c c}
            \toprule
            \mrow[l]{Training \\Setup} & \mrow{Accuracy\\(MSE)} & \mrow{Throughput\\(samples/sec)}\\
            \cmidrule(lr){1-3}
            MovieLens & 0.794\(\pm\)0.020                          & \textbf{22967.02\(\pm\)2468.72} \\
            IMDB      & \multicolumn{2}{c}{used as auxiliary data} \\
            Combined  & 0.788\(\pm\)0.015                          & 20843.26\(\pm\)2194.09 \\
            SplitNN   & \textbf{0.785\(\pm\)0.015}                 & 19496.39\(\pm\)1900.94 \\
            \bottomrule
        \end{tabular}
    }
    \subtable[Statistical Test Results]{
        \footnotesize
        \setlength{\tabcolsep}{.5ex}
        \begin{tabular}{lll}
            \toprule
            Workload                                     & Metric                      & Test Results\(^{1}\) \\
            \cmidrule(lr){1-3}
            \multirow{6}{*}[-.5ex]{Year Prediction}      & \multirow{3}{*}{Accuracy}   &
            \(p(\text{SplitNN},\text{Equal},\text{FMA})=\langle 1.00, 7.22\times10^{-8}, 7.26\times10^{-8}\rangle\) \\ &&
            \(p(\text{SplitNN},\text{Equal},\text{MSD})=\langle 1.69\times10^{-10}, 7.42\times 10^{-10}, 1.00\rangle\) \\ &&
            \(p(\text{SplitNN},\text{Equal},\text{Combined})=\langle 5.41\times 10^{-2}, 8.09\times 10^{-1}, 1.35\times 10^{-1}\rangle\) \\
            \cmidrule(lr){2-3}
                                                         & \multirow{3}{*}{Throughput} &
            \(p(\text{SplitNN},\text{Equal},\text{FMA})=\langle 1.85\times 10^{-7}, 3.43\times 10^{-6}, 1.00\rangle\) \\ &&
            \(p(\text{SplitNN},\text{Equal},\text{MSD})=\langle 1.67\times 10^{-12}, 3.38\times 10^{-12}, 1.00\rangle\) \\ &&
            \(p(\text{SplitNN},\text{Equal},\text{Combined})=\langle 1.96\times 10^{-5}, 1.89\times 10^{-3}, 9.98\times 10^{-1}\rangle\) \\
            \cmidrule(lr){1-3}
            \multirow{4}{*}[-.5ex]{Movie Recommendation} & \multirow{2}{*}{Accuracy}   &
            \(p(\text{SplitNN},\text{Equal},\text{MovieLens})=\langle 1.14\times 10^{-3}, 6.23\times 10^{-1}, 3.75\times 10^{-1}\rangle\) \\ &&
            \(p(\text{SplitNN},\text{Equal},\text{Combined})=\langle 4.85\times 10^{-4}, 9.81\times 10^{-1}, 1.78\times 10^{-2}\rangle\) \\
            \cmidrule(lr){2-3}
                                                         & \multirow{2}{*}{Throughput} &
            \(p(\text{SplitNN},\text{Equal},\text{MovieLens})=\langle 8.19\times 10^{-9}, 3.19\times 10^{-6}, 1.00\rangle\) \\ &&
            \(p(\text{SplitNN},\text{Equal},\text{Combined})=\langle 5.72\times 10^{-7}, 9.24\times 10^{-2}, 9.08\times 10^{-1}\rangle\) \\
            \bottomrule
            \multicolumn{3}{p{0.9\textwidth}}{
                \(^{1}\) For Accuracy of the year prediction task, rope=\(\pm\)0.1. For Accuracy of the movie recommendation task, rope=\(\pm\)0.01. For throughput of both tasks, rope=\(\pm\)1000.
            }
        \end{tabular}
    }
\end{table}

Table~\ref{tab:experiment-vert-baseline} shows the accuracy results of both tasks. Similar to horizontal FL, we observe that the federated setup achieves similar or higher accuracy performance compared to training with the combined dataset. However, the year prediction task shows that the performance of vertical FL is not always better than training with a single dataset.

Unlike horizontal FL, as samples from the different datasets need to be linked with each other before being fed into the model, and each set of linked samples are only counted as a single sample, the end-to-end throughput in the combined setup is lower than then single-detest setups. In the SplitNN setup, the communication between models further lowers the throughput.

\textit{Summary:} Vertical FL algorithm results in similar accuracy compared to the combined setup, but achieves lower throughput.

\subsection{Vertical Federated Learning\@: Homomorphic Encryption}\label{sub:homomorphic_encryption_in_vertical_federated_learning}

\paragraph{Experimental Setup}
Due to the model heterogeneity of involved parties, SMC via secret sharing cannot simply be applied to encrypt the data in SplitNN\@. Therefore, we performed trials with homomorphic encryption on the recommendation task to test its performance.

\paragraph{Results and Analysis}

Table~\ref{tab:experiment-vert-encryption} shows the result of HE compared to the baseline. In our experiment, the average spent time of each batch is around 206 seconds, and since there are 7834 batches per epoch, and every batch needs to be encrypted independently to preserve security, the estimated total training time is about 403 hours, which is unacceptable in real practice. Compared with the non-federated setting, the huge increase of training time is mainly incurred by the homomorphic encryption, while the communication time is negligible. Finding a efficient encryption scheme for HE is essential for the development of practical vertical FL applications.

\begin{table}[htbp]
    \centering
    \caption{Test MSE and training time of the movie recommendation task}\label{tab:experiment-vert-encryption}
    \begin{tabular}{l c c}
        \toprule
        Train Setup     & \mrow{Convergence \\Time (rounds)} & \mrow{Throughput\\(samples/sec)} \\
        \cmidrule(lr){1-3}
        Combined        & 29.50\(\pm\)0.76  & 20843.26\(\pm\)2194.09 \\
        SplitNN         & 29.08\(\pm\)0.76  & 19496.39\(\pm\)1900.94 \\
        SplitNN with HE & N/A               & about 0.61 \\
        \bottomrule
    \end{tabular}
\end{table}